\def\year{2022}\relax
\documentclass[letterpaper]{article} % DO NOT CHANGE THIS
\usepackage{aaai22}  % DO NOT CHANGE THIS
\usepackage{times}  % DO NOT CHANGE THIS
\usepackage{helvet}  % DO NOT CHANGE THIS
\usepackage{courier}  % DO NOT CHANGE THIS
\usepackage[hyphens]{url}  % DO NOT CHANGE THIS
\usepackage{graphicx} % DO NOT CHANGE THIS
\usepackage{natbib}  % DO NOT CHANGE THIS AND DO NOT ADD ANY OPTIONS TO IT
\usepackage{caption} % DO NOT CHANGE THIS AND DO NOT ADD ANY OPTIONS TO IT

\DeclareCaptionStyle{ruled}{labelfont=normalfont,labelsep=colon,strut=off} % DO NOT CHANGE THIS
\usepackage{times}
\usepackage{epsfig}
\usepackage{graphicx}
\usepackage{amsmath}
\usepackage{amssymb}
\usepackage{bm}
\usepackage{bbm}
\usepackage{times}
\usepackage{epsfig}
\usepackage{graphicx}
\usepackage{amsmath}
\usepackage{amssymb}
\usepackage{subfigure}
\usepackage{color}
\usepackage{bbding}
\usepackage{caption}
\usepackage{bbding}
\usepackage{booktabs}
\usepackage{makecell}
\usepackage{algorithm}
\usepackage{algorithmic}

\usepackage{newfloat}
\usepackage{listings}
\floatstyle{ruled}
\newfloat{listing}{tb}{lst}{}
\floatname{listing}{Listing}
\title{Visual Sound Localization in the Wild by Cross-Modal Interference Erasing}
\author {
    Xian Liu\textsuperscript{\rm 1,2}\thanks{Equal contribution. $^\sharp$ Corresponding author.},
    Rui Qian\textsuperscript{\rm 1,3}\footnotemark[1] ,
    Hang Zhou\textsuperscript{\rm 1}\footnotemark[1] ,
    Di Hu\textsuperscript{\rm 4},
    Weiyao Lin\textsuperscript{\rm 3}, 
    Ziwei Liu\textsuperscript{\rm 5},\\
    Bolei Zhou\textsuperscript{\rm 1},
    Xiaowei Zhou\textsuperscript{\rm 2$\sharp$}
}
\affiliations {
    \textsuperscript{\rm 1} The Chinese University of Hong Kong,
    \textsuperscript{\rm 2} Zhejiang University,
    \textsuperscript{\rm 3} Shanghai Jiao Tong University, \\
    \textsuperscript{\rm 4} Gaoling School of Artificial Intelligence, Renmin University of China,
    \textsuperscript{\rm 5} S-Lab,  Nanyang Technological University

    alvinliu@ie.cuhk.edu.hk, qr021@ie.cuhk.edu.hk, zhouhang@link.cuhk.edu.hk
}
\usepackage{bibentry}
\begin{document}

\maketitle

%%%%%%%%% ABSTRACT
\begin{abstract}
The task of audio-visual sound source localization has been well studied under constrained scenes, where the audio recordings are clean. However, in real-world scenarios, audios are usually contaminated by off-screen sound and background noise. They will interfere with the procedure of identifying desired sources and building visual-sound connections, making previous studies non-applicable. In this work, we propose the \textbf{Interference Eraser (IEr)} framework, which tackles the problem of audio-visual sound source localization in the wild. The key idea is to eliminate the interference by redefining and carving discriminative audio representations. Specifically, we observe that the previous practice of learning only a single audio representation is insufficient due to the additive nature of audio signals. We thus extend the audio representation with our Audio-Instance-Identifier module, which clearly distinguishes sounding instances when audio signals of different volumes are unevenly mixed. Then we erase the influence of the audible but off-screen sounds and the silent but visible objects by a Cross-modal Referrer module with cross-modality distillation. Quantitative and qualitative evaluations demonstrate that our proposed framework achieves superior results on sound localization tasks, especially under real-world scenarios. Code is available at \url{https://github.com/alvinliu0/Visual-Sound-Localization-in-the-Wild}.
\end{abstract}

%%%%%%%%% BODY TEXT

\section{Introduction}
\begin{figure}
    \centering
    \setlength{\abovecaptionskip}{2mm}
    \includegraphics[width=\linewidth]{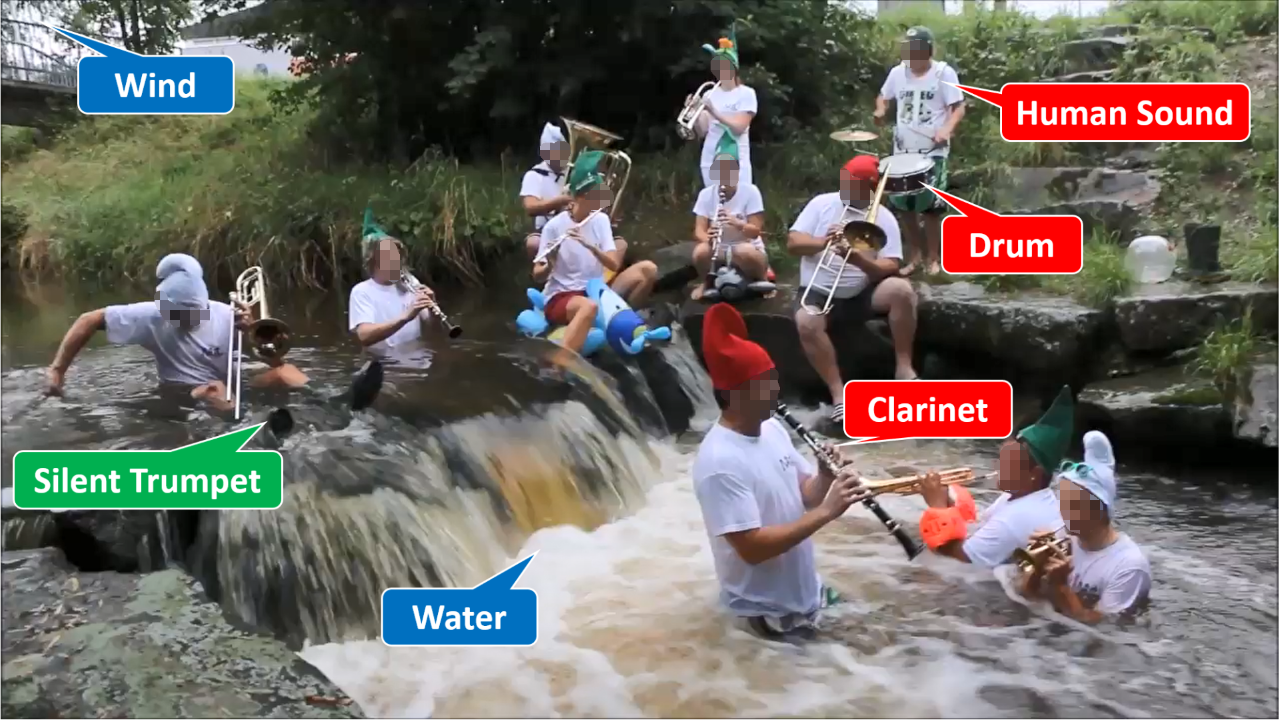}
    \caption{\textbf{An example of an in-the-wild scene\protect\footnotemark.} It contains sounding clarinet, drum, human, etc. (red); visible but silent trumpet (green); background sound of water and wind (blue). Given an image and its corresponding audio as \textbf{input}, we aim to localize objects that produce sound in the image.}
    \label{fig:cover}
\end{figure}

 Humans can grasp the relevance between audio and visual information by leveraging their natural correspondences~\cite{proulx2014multisensory,stein1993merging}. Even for an in-the-wild scenario as shown in Fig.~\ref{fig:cover}, humans are able to distinguish the sounding objects despite the background sound and noises. In order for machines to achieve human-like multi-modality perception, researchers have explored the task of visual sound source localization~\cite{zhao2018sound,senocak2018learning,hu2020discriminative,qian2020multiple}, which aims at localizing the objects that produce sound in an image given its corresponding audio clip. 

\footnotetext{ \url{https://www.youtube.com/watch?v=3dXbFEOtIbs.}}

With the recent development of neural networks, studies have firstly been made on learning audio-visual correspondences towards associating sound and visual objects. They propose to optimize the similarity between audio and visual features according to spatial-level correspondences or temporal synchronization. Such pipeline effectively detects sound sources in scenarios with single sound source~\cite{arandjelovic2017look,arandjelovic2018objects,owens2018audio,senocak2018learning}. However, models learned under such settings cannot generalize well to mixed sound scenes. To further improve the model's capacity, fine-grained audio-visual components are leveraged to establish sound-object associations for multiple sound source localization. \citet{qian2020multiple} propose to solve the problem in a coarse-to-fine manner. But it relies on audio class labels, which are sometimes inaccessible. \citet{hu2020discriminative} design a two-stage pipeline to transfer single-sound prior to multi-sound cases and suppress the influence of silent objects. Nevertheless, they care less about the interference in audios, thus confining the applicable scenarios to clean environments.

The interference in audio is caused by the additive nature of audio signals. When the audio clip is a mixture of different sources and noise, several challenges occur: 1) In the localization task, sound sources of different volumes should be evenly identified. However, machines tend to distinguish only a few dominant sounds from a mixture as demonstrated by~\citet{phan2017makes,adavanne2018sound}. Thus the normal practice of previous studies~\cite{arandjelovic2017look,arandjelovic2018objects,qian2020multiple,hu2020discriminative} to learn a single audio embedding is not enough for such complicated scenarios. More distinguishable representations are needed. 2) The undesired off-screen sound and background noise will be evenly taken into account during audio encoding, which might compromise the procedure of audio-visual modality matching.

In this paper, we propose a label-free method targeting in-the-wild visual sound localization called \textbf{Interference Eraser (IEr)}. The key idea is to \emph{eliminate the interference by redefining and carving discriminative audio representations}. We regard volume differences between mixed sound sources, audible but off-screen sounds, and visible but silent objects as \emph{interference} in the task of visual sound localization. 
In order to erase them, two questions need to be addressed: 1) How to identify audio instances regardless of their volume? 2) How to match the visible and sounding instances from the two modalities? 

Specifically, we first build audio and visual \emph{prototypes} in the feature space through learning the traditional single-source audio-visual correspondences~\cite{arandjelovic2018objects}. The prototypes are the centroid features of each clustered category, which are responsible for recognizing individual instances in the absence of label information. For the case of mixed sound sources, we design an \textbf{Audio-Instance-Identifier} module to evenly identify all sounding instances by learning a latent step for each prototype to compensate for the mixture's differences in volumes. In this way, we expand the audio representation from a single feature to a set of features, each taking care of one sounding category. At the second stage, we propose the \textbf{Cross-modal Referrer} that particularly targets the off-screen interference. 
The confidence score of all cluster-classes are transformed to class distributions for both domains, and we particularly focus on erasing the interference on the distributions by cross-referring the two modalities. Finally, the predicted distributions on all clustered categories from both modalities serve as soft targets for each other to perform cross distillation. 

Our contributions are summarized as follows:
\begin{itemize}
    \item We design the Audio-Instance-Identifier module which erases the influence of sound volume in uneven audio mixtures for discriminative audio perception.
    \item We devise the Cross-modal Referrer module to symmetrically erase the interference of silent but visible objects and audible but off-screen sounds.
    \item Through cross-modal distillation learning, our proposed \textbf{Interference Eraser} remarkably outperforms the state-of-the-art methods in terms of sound localization for both synthetic and in-the-wild audio-visual scenes. 
\end{itemize}

\section{Related Work}
\setcounter{secnumdepth}{2}
\noindent\textbf{Audio-Visual Correspondence Learning.}
\label{related}
The correspondence between co-occurring modalities, \emph{e.g.}, audio and vision, provides natural and expressive self-supervision for representation learning in multi-modal scenarios~\cite{korbar2018cooperative,aytar2016soundnet,tian2018audio,Zhou_2019_ICCV,zhou2019talking,zhou2021pose}. Recent works~\cite{alwassel2019self,patrick2020multi} use temporal consistency as a self-supervisory signal to facilitate audio-visual learning. Typically, \citet{arandjelovic2018objects} calculate the similarity between global audio embedding and visual features of each spatial grid to generate audio-visual correlation maps. \citet{zhao2018sound,zhao2019sound} employ audio-visual consistency to align channel-wise cross-modal distributions, then leverage the learned representation for sound source separation. However, most previous works are performed in single-source settings. 
In real-world scenarios with multiple sources, as proved in previous works on sound separation~\cite{tzinis2021improving,gao2019co,gan2020music,pu2017audio}, the audio prediction could be quite unbalanced, which affects the mapping between audio and visual modality. In this work, we focus on in-the-wild scenes.

\begin{figure*}
    \centering
    \setlength{\abovecaptionskip}{2mm}
    \includegraphics[width=\linewidth]{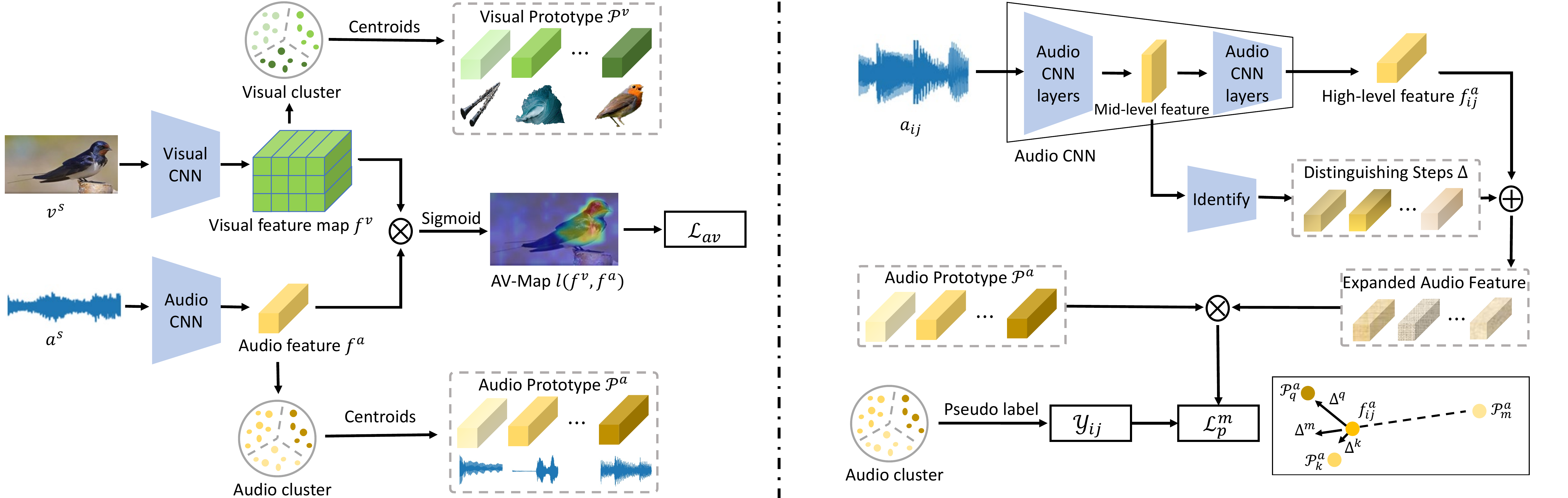}
    \caption{\textbf{An illustration of the first-stage learning.} The \textbf{left} presents audio-visual prototype extraction. We calculate the cosine similarity between the audio embedding and the visual feature at each position to generate the localization map, and cluster audio-visual features into centroids as prototypes. The \textbf{right} shows Audio-Instance-Identifier. Given mixed audio, we predict distinguishing-steps to generate expanded audio features. As shown in the corner, it predicts a small/big step for dominant/non-dominant sound to approach all sounding prototypes, and a reverse step to repel from non-existent sounds' prototypes.}
    \label{fig:stage1}
\end{figure*}

\noindent\textbf{Visual Sound Source Localization.}
Visually localizing sound source aims to figure out specific sounding area in the visual scene given a clip of audio, and  can be achieved in various ways~\cite{arandjelovic2018objects,chen2021localizing,lin2021unsupervised,zhou2020sep,xu2021visually,tian2021cyclic,tian2021can,afouras2020self}. For example, \citet{arandjelovic2018objects,owens2018audio} perform sound localization by measuring similarity between audio embedding and visual features on spatial area. ~\citet{senocak2018learning} develop audio-guided attention mechanisms to refine audio-related visual features and improve cross-modal alignment. In complex scenes, \citet{zhao2018sound} introduce sound separation as a pretext task to perform visual grounding through channel correlation. 
Particularly, \citet{qian2020multiple} target multi-sound scenes by adopting category prior to associate one-to-one sound-object pairs. One most recent work~\cite{hu2020discriminative} proposes to discriminatively localize sounding objects via category-level audio-visual distribution alignment. Specifically, it first trains audio-visual correspondence in single-sound scenes, then \emph{directly} transfers to mixed sound cases for audio-visual matching. However, the quite unbalanced audio mixture and off-screen sound in multi-source scenarios impose great challenges to previous works. Alternatively, we develop the Interference Eraser to achieve robust mixed sound perception and filter out interference. 

\section{Method}
\begin{figure*}[t]
    \centering
    \setlength{\abovecaptionskip}{2mm}
    \includegraphics[width=\linewidth]{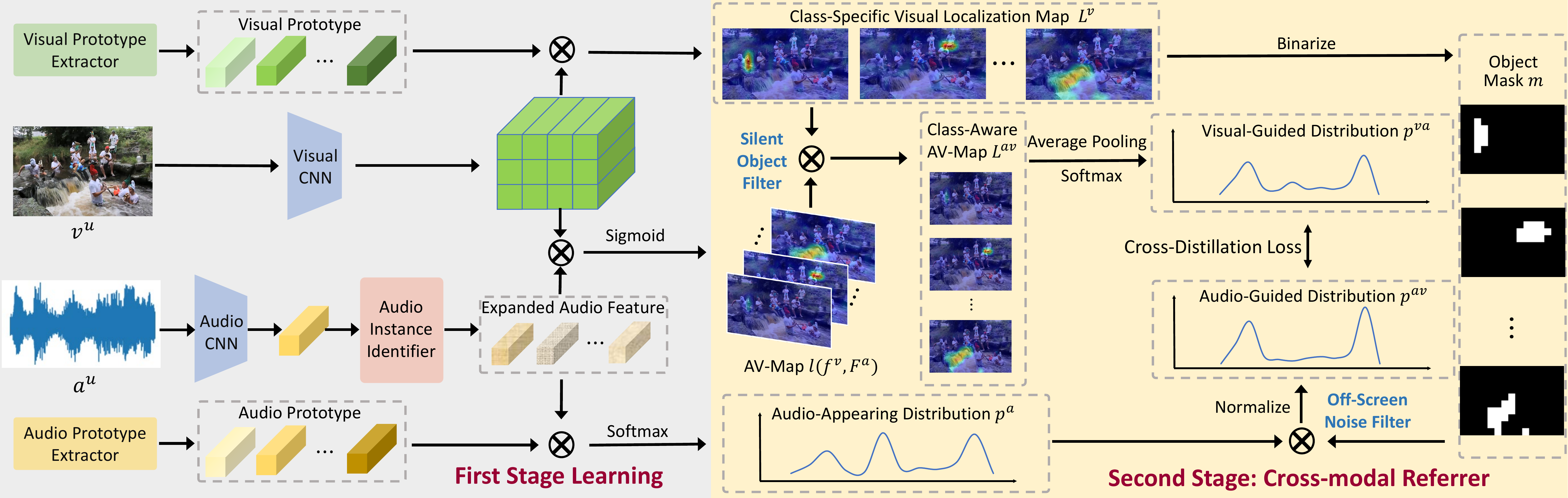}
    \caption{\textbf{An overview of the Interference Eraser (IEr) framework.} The \textbf{left} presents the first stage, where we train on single-sound data to establish audio-visual prototypes, and then utilize Audio-Instance-Identifier to facilitate class-specific audio features. The \textbf{right} presents the second stage, where the silent object filter suppresses visual interference by referring to balanced audio distribution, and the off-screen noise filter eliminates audio interference by referring to binary visual masks. A cross-distillation loss helps to achieve robust sound localization in-the-wild. The class-aware AV-Maps $L^{av}$ are the final output in inference.}
    \label{fig:framework}
\end{figure*}

In this section, we introduce the details of our \textbf{Interference Eraser (IEr)} for label-free visual sound localization. The whole pipeline is illustrated in Fig.~\ref{fig:framework}. 
Below we first illustrate the first-stage learning: audio-visual prototype extraction by single-source audio-visual learning (Sec.~\ref{sec:3.1}), and the Audio-Instance-Identifier with a curriculum learning scheme for mixed-audio instance discrimination (Sec.~\ref{sec:3.2}). Then, we introduce the second-stage learning, \emph{i.e.}, Cross-modal Referrer (Sec.~\ref{sec:3.3}) to 
perform in-the-wild scene training and distribution matching for robust sound localization.

Notably, the training videos are denoted as $\mathcal{X} = \{(a_i, v_i) | i = 1, 2, ..., N\}$, where $N$ is the number of videos and $(a_i, v_i)$ is the $i$-th audio-visual pair. $\mathcal{X}$ can be divided into two disjoint subsets: $\mathcal{X}^s = \{(a_i^s, v_i^s) | i = 1, 2, ..., N^s\}$ consisting of $N^s$ single source videos and $\mathcal{X}^u = \{(a_i^u, v_i^u) | i = 1, 2, ..., N^u\}$ consisting of $N^u$ unconstrained in-the-wild videos. The first stage training is conducted on $\mathcal{X}^s$, and the second stage is conducted on $\mathcal{X}^u$.

\subsection{Audio-Visual Prototype Extraction}
\label{sec:3.1}

Towards localizing sound-making objects, the first step is to associate the visual appearance of objects with their sound. While visual instances can be divided spatially, audio signals are naturally mixed together in real-world scenarios. 
Moreover, no label information can be leveraged in our setting. In order to discriminate different instances from a mixture, representations of each possible sounding category need to be identified.

To this end, our first step is to perform the traditional instance-level feature correspondence learning on the single source subset $\mathcal{X}^s$~\cite{arandjelovic2018objects,zhao2018sound} and find the latent representatives for each class in the feature space, namely the \emph{prototypes} in the same way as~\cite{hu2020discriminative}. 

\noindent\textbf{Single Source Audio-Visual Learning.}
As illustrated in Fig.~\ref{fig:stage1} left, we employ the variants of ResNet-18~\cite{he2016deep} as visual and audio backbones to extract visual feature map $f^v\in \mathbb{R}^{H\times W \times C}$ and the audio feature $f^a \in \mathbb{R}^{C}$ from $\mathcal{X}^s$. The cosine similarity between $f^a$ and each positional feature $f^{v_{(x,y)}}$ of $f^v$ are used to present an audio-visual association map (localization map) $l(f^v, f^a) \in \mathbb{R}^{H\times W}$. The training is through contrastively positive and negative pair sampling:
\begin{align}
    \mathcal{L}_{av} &= \mathcal{L}_{bce}(\text{GMP}(l(f^a_i,f^v_j)),\delta),
    \label{avloss}
\end{align}
where \text{GMP} denotes Global Max Pooling, $\mathcal{L}_{bce}$ is binary cross-entropy loss function and $\delta$ indicates whether $f^a_i$ and $f^v_j$ are extracted from the same video, \emph{i.e.}, $\delta=1$ when $i=j$, otherwise $\delta=0$. 

\noindent\textbf{Embedding Prototypes.}
As visual appearances of different objects are comparatively easier to learn, the categories can be roughly defined through deep clustering~\cite{caron2018deep,alwassel2019self} the visual object features extracted from $\mathcal{X}^s$. We manually set a large cluster number and finally settle a total number of $K$ categories. The visual \emph{prototype} ${\mathcal{P}^v_j} \in  \mathbb{R}^{C}$ is defined as the centroid for the $j$-th class in the feature space. All visual prototypes can be represented as $\mathcal{P}^v \in \mathbb{R}^{K\times C}$. 

On the other hand, the pseudo-class labels can be automatically assigned to the video's accompany audios. Thus the audio prototypes $\mathcal{P}^a \in \mathbb{R}^{K\times C}$ can be identified in the same way as visual ones. In this way,  a pseudo-class $k$ can be assigned to each single source audio-visual data pair $(a^{s_k}_i, v^{s_k}_i)$ in $\mathcal{X}^s$,

\subsection{Audio-Instance-Identifier}
\label{sec:3.2}
As stated before, localizing sounding objects in multi-source scenes requires identifying each object.
Even with the prototypes learned, one embedded feature from a mixture of audios can not be similar to all prototypes, thus limits the feature's capacity to identify all instances.
Particularly in an audio mixture with different volumes, the dominant sound might severely interfere with the audio discrimination. 

To accurately identify different sounding components, we propose the Audio-Instance-Identifier module which expands the audio representation. The key is to learn a \emph{Distinguishing-step} for each prototype in a curriculum manner as shown in the right part of Fig.~\ref{fig:stage1}.

\noindent\textbf{Learning the Distinguishing-step.}
 \label{sec:3.2.2}
Inspired by research in sound source separation~\cite{zhao2018sound,gao2019co}, we propose to manually mix single sources together as input and aim at recognizing them individually.
We denote the pseudo label for an single audio clip $a^{s_k}_i$ as $\mathcal{Y}_i \in \mathbb{R}^{K}$ whose $k$-th element $\mathcal{Y}^k_i = 1$ and others $\mathcal{Y}^n_i = 0$ when $n \neq k$.
By mixing the $i$-th audio clip $a^{s_k}_i$ with pseudo-class $k$ and the $j$-th clip $a^{s_q}_j$ with pseudo-class $q (q \neq k)$ together, a mixture $a_{ij}$ with pseudo label $\mathcal{Y}_{ij} = \mathcal{Y}_{i} + \mathcal{Y}_{j}$ can be rendered. 

In order to erase the interference of volume difference and identify all sounding categories, we propose to learn a set of \emph{distinguishing-steps} $\Delta^n(a) \in \mathbb{R}^{K\times C}$ towards each prototype $\mathcal{P}^a_n$ from the input audio. The motivation is to compensate for the amplitude information in the feature space. For example, as shown in Fig.~\ref{fig:stage1}, in the two-mix case listed above, when the learned feature $f^a_{ij}$ is too close to the dominant $\mathcal{P}^a_k$ but far away from $\mathcal{P}^a_q$, the desired $\Delta^q(a_{ij})$ should be a large step towards $\mathcal{P}^a_q$ while $\Delta^k(a_{ij})$ is a smaller step towards $\mathcal{P}^a_k$. 
We can formulate a binary classification problem for each pseudo-class:
\begin{align}
    \mathcal{L}^m_p &= \frac{1}{K}\sum^{K}_{n=1} \mathcal{L}_{bce}(sim({\mathcal{P}^a_n}, f^a_{ij} + \Delta^n(a)), \mathcal{Y}^n_{ij}).
    \label{eq:pro2}
\end{align}
where $sim(f_1, f_2)$ is the cosine similarity function defined as $sim(f_1, f_2) = \frac{\bm{f}_1\cdot \bm{f}_2}{|\bm{f}_1||\bm{f}_2|}$.
It can be seen that this distinguishing-step also helps to repel from the non-existent audio instances' prototypes by predicting a step to the reverse direction of $\mathcal{P}^a_m  (m \neq q,k)$ as shown in Fig.~\ref{fig:stage1}.

Particularly, distinguishing-step is not directly learned from high-level feature $f^a_{ij}$. It is verified that mid-level features can capture more fine-grained information~\cite{yosinski2014transferable,lin2017feature,zou2020revisiting}, thus we employ the weighted combination of linearly transformed mid-level features (shown in Fig.~\ref{fig:stage1}) to make prediction\footnote{Please refer to Supple. for details of distinguishing-step.}.

\noindent\textbf{Curriculum Learning Strategy.}
\label{sec:curriculum}
We develop a curriculum learning strategy for the training process of the Audio-Instance-Identifier. Detailedly, we choose to alternatively train on single-source data without the discriminating-step and mixed sound with $\mathcal{L}^m_p$ (Eq.~\ref{eq:pro2}). In each epoch, a proportion of $p < 1$ samples are mixed ones. $p$ is initialized as 0.5 and gradually increases to 0.9 at the end of training. While Eq.~\ref{eq:pro2} shows only a two-mix case, it can be expanded to cases with 3 or more mixtures. The number of mixtures is set as $2$ at first and gradually increases to $4$ during the training process. As the audio backbone is keeping updated, the audio prototypes are re-computed at the beginning of each epoch.

\begin{table*}
	\centering
	\renewcommand{\arraystretch}{1.1}
  \begin{tabular}{l|cc|cc|cc|cc|cc|cc}
    \toprule[1.2pt]
    Scenario & \multicolumn{4}{c|}{Single Sound Scene (a)} & \multicolumn{8}{c}{General in-the-Wild Scene (b)} \\
    \hline
    Dataset & \multicolumn{2}{c|}{MUSIC} &
    \multicolumn{2}{c|}{VGGSound} &
    \multicolumn{2}{c|}{MUSIC-Syn.} & \multicolumn{2}{c|}{MUSIC-Duet} & \multicolumn{2}{c|}{MUSIC-Un.} & \multicolumn{2}{c}{VGG-Un.} \\\hline
    Methods\textbackslash Metrics & IoU & AUC & IoU & AUC & CIoU & AUC & CIoU & AUC & CIoU & AUC & CIoU & AUC\\
    \hline 
     Object-that-sound & 26.1 & 35.8 & 48.4 & 46.1 & 3.7  & 10.2 & 13.2  & 18.3 & 0.1 & 6.8 & 7.8 & 15.1 \\    
     Sound-of-pixel & 40.5 & 43.3 & 42.5 & 45.1  & 8.1  & 11.8  & 16.8 & 16.8 & 7.5 & 11.6 & 7.9 & 14.4 \\
     DSOL  & 51.4 & 43.6 & 49.3 & 45.8  & 32.3  & 23.5 & 30.2  & 22.1 & 3.2 & 7.3 & 8.1 & 12.2 \\
     \hline
     Interference Eraser & \textbf{53.9} & \textbf{50.7} & \textbf{51.3} & \textbf{46.9}  & \textbf{47.6}  & \textbf{29.8} & \textbf{52.9}   & \textbf{33.8} & \textbf{15.6} & \textbf{15.3} & \textbf{12.8} & \textbf{17.6} \\
     \bottomrule[1.2pt]
  \end{tabular}
  \caption{\textbf{Visual Sound Localization results in (a) Single Sound Scene:} MUSIC, VGGSound; \textbf{(b) General in-the-Wild Scene:} MUSIC-Syn., MUSIC-Duet, MUSIC-Un., VGG-Un. We compare \textbf{IEr} against recent SOTA methods. The IoU reported in Table(a) is IoU@0.5, and the CIoU reported in Table(b) is CIoU@0.3. The results of \cite{arandjelovic2018objects,zhao2018sound,hu2020discriminative} on MUSIC, MUSIC-Syn. and MUSIC-Duet are reported from~\cite{hu2020discriminative}.}
  \label{tbl:syn}
\end{table*}

\subsection{Cross-modal Referrer}
\label{sec:3.3}
At the second stage, we move to real-world multi-source scenarios, where the training and inference are carried out on $\mathcal{X}^u$. Normally, the localization map on general scenes can be calculated in the same way as learning audio-visual correspondences, that is, to directly encode the audio feature $f^a$ from a mixed audio $a^u$, and compute cosine similarities $l(f^v, f^a)$ on the visual feature map $f^v$ (Sec.~\ref{sec:3.1}). However, this operation will suffer from the same 
uneven audio feature encoding problem as stated in Sec.~\ref{sec:3.2}. 

Thus, we propose the Cross-modal Referrer as shown in Fig~\ref{fig:framework}. It first identifies the audio-visual instances, and then 
predicts the class-aware audio-visual localization maps where visual interference is erased by referring to our updated audio representation. 
Afterwards, we leverage the acquired localization map to erase the audio interference and compute the category distributions of both audio and visual domains. Finally, a cross-modal distillation loss is used to facilitate better cross-modality matching. 

\noindent\textbf{Class-Aware Audio-Visual Localization Maps.}
Given an input image with the visual prototypes $\mathcal{P}^v$, it is easy for us to find all potential sound-making objects by computing the visual localization map for the category $k$ as $L^v_k = l(f^v, \mathcal{P}^v_k), L^v_k \in \mathbb{R}^{H \times W}$. However, within the $L^v$s, there might be responses on silent but visible object's map $L^v_m$, thus we refer to the audio domain.

We suppose the audio clip $a^u$ accompanies $v^u$ is a mixture of sources including an off-screen sound $a_j$.  Its feature $f^a$ together with its \emph{distinguishing-steps} $\boldsymbol{\Delta}(a) = [\Delta^1(a), \dots, \Delta^K(a)]$ for each pseudo class can be computed as illustrated in Sec.~\ref{sec:3.2.2}. The audio feature can be expanded to a set of \emph{class-specific} features ${F}^a_k = f^a + {\Delta}^k(a)$. 

Then we can predict the \emph{class-aware audio-visual localization maps} by associating the class-specific audio features ${F}^a$ with the visual feature map $f^v$ again, then mask it by the class-specific visual localization maps $L^v$. It is computed as $L^{av}_k = l(f^v, F^a_k) \cdot L^v_k$ for each class $k$, where $\cdot$ is the element-wise matrix multiplication. As there would be little responses on the silent object $m$ for $l(f^v, F^a_m) \cdot L^v_m$, the interference of $m$ is thus suppressed. This whole operation is marked as the \textbf{Silent Object Filter} on Fig.~\ref{fig:framework}. Note that these class-aware audio-visual localization maps (AVMaps) are also the desired output of \textbf{IEr} system during inference.

\noindent\textbf{Acquiring Audio-Visual Distribution.}
Besides acquiring the audio-visual localization maps (AVMaps), we can also predict the on-screen objects' distributions on the pseudo-classes. The way is to perform global average pooling (GAP) on each AVMap as a rough accumulation of the class's response. Then use a softmax function to regularize all responses on each class. This visual-guided distribution can be written as:
\begin{align}
    \text{p}^{va} = \text{softmax}([\text{GAP}(L^{av}_1), \dots, \text{GAP}(L^{av}_K)]).
\end{align}

On the other hand, we can also compute the audio-appearing probability for each pseudo-class as cosine similarity between the audio prototype and class-specific audio features $\text{p}^a_k = sim({\mathcal{P}^a_k}, F^a_k)$. However, the off-screen sound $a_j$ might interfere with the final on-screen sound distribution prediction. To erase this interference, we refer to the help of visual localization maps $L^v$. Specifically, we first binarize each visual localization map $L^v_k$ to a class-specific binary mask $m_K$ by setting a threshold. This is to suppress noisy peaks in $L^v_k$. 
Then we leverage the global average-pooled binary mask $\text{GAP}(m_k)$ to accounts for the portion of this instance in the visual domain. Thus the visual-associated audio-appearing probability on class $k$ can be written as:
\begin{align}
{\text{p}}^{av}_k  =  \frac{\text{GAP}(m_k) \cdot \text{p}^a_k}{\sum_{k=1}^K\text{GAP}(m_k) \cdot \text{p}^a_k}.
\end{align}
After this operation, the influence of off-screen sound $a_j$ can be erased, as the reference visual contribution of its corresponding class would be subtle. This whole operation is marked as the \textbf{Off-screen Noise Filter} on Fig.~\ref{fig:framework}.
The audio-guided class distribution can then be computed as:
\begin{align}
    {\text{p}}^{av} = \text{softmax}([\text{p}^{av}_1, \dots, \text{p}^{av}_K]).
\end{align}

Notably, the visual size of objects does not necessarily connect with the volume of their sound. Thus this cross-modal referring operation on the audio distribution not only erase off-screen sounds but also modulate the audio subject's probability according to its visual size, which would be beneficial to cross-modality distribution matching.

\noindent\textbf{Cross-Modal Distillation.}
Finally, given the class distributions $\text{p}^{va}$ and ${\text{p}}^{av}$, which are guided by the visual and audio modalities respectively, we can regard one of them as soft targets in the formulation of knowledge distillation~\cite{44873}. Substantially, it is to leverage the Kullback–Leibler divergence $\mathcal{D}_{KL}$ for similarity measurement between the two distributions. This additional training objective for in-the-wild scene is:
\begin{align}
     \mathcal{L}_u&= \frac{1}{2}\mathcal{D}_{KL} (\text{p}^{va}||{\text{p}}^{av}) + \frac{1}{2}\mathcal{D}_{KL} ({\text{p}}^{av} || \text{p}^{va}).
    \label{loss-kld}
\end{align}

\section{Experiments}

\subsection{Dataset}

\noindent \textbf{MUSIC (Synthetic)} MUSIC dataset~\cite{zhao2018sound} covers 11 types of instruments. Since some YouTube videos have been removed, we collect 489 solo and 141 duet videos. Following~\cite{hu2020discriminative}, we employ half of the solo data to form $\mathcal{X}^s$, with the other half to synthesize $\mathcal{X}^u$. Specifically, we use the first five/two videos of each category in solo/duet to evaluate localization performance and employ the rest videos to generate MUSIC-Synthetic~\cite{hu2020discriminative} as well as MUSIC-Unconstrained. 

\noindent \textbf{VGGSound} VGGSound dataset~\cite{vedaldi2020vggsound} consists of more than 210k single-sound videos, covering 310 categories. For better evaluation, we filter out a subset of 98 categories with specific sounding objects that can be localized. Similarly, we use half of the data for single-sound and the other half for synthesizing VGG-Synthetic. We finally obtain 28,756 videos for training and 2,787 for evaluation. This subset is very suitable for class-aware sound localization and also provides diverse challenging scenarios. 

\noindent \textbf{Unconstrained-Synthetic} Since most samples in two datasets do not simultaneously contain the aforementioned interference, they fail to fully reflect the difficulty of in-the-wild scenarios. To better validate sound localization in general scenes, we establish two challenging datasets called MUSIC-Unconstrained and VGG-Unconstrained. For each synthetic audio-visual pair, the image contains two sounding and two silent objects, and the audio is a mixture of on-screen and randomly chosen off-screen audio clips. In this way, the Unconstrained dataset simulates in-the-wild scenes with visible but silent objects and off-screen noise.

\begin{figure*}[t]
    \subfigure[Sound-of-pixel~\cite{zhao2018sound}]{
    \includegraphics[width=0.117\linewidth]{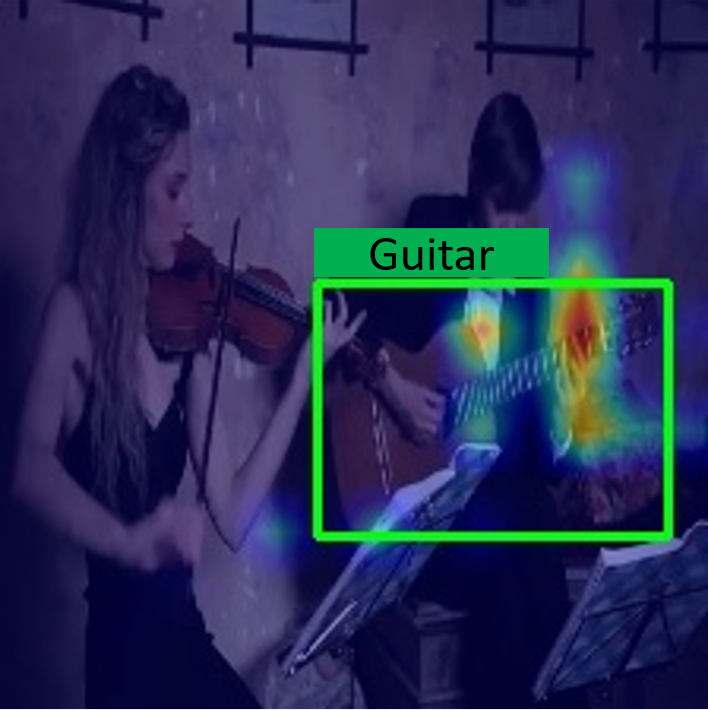}
    \includegraphics[width=0.117\linewidth]{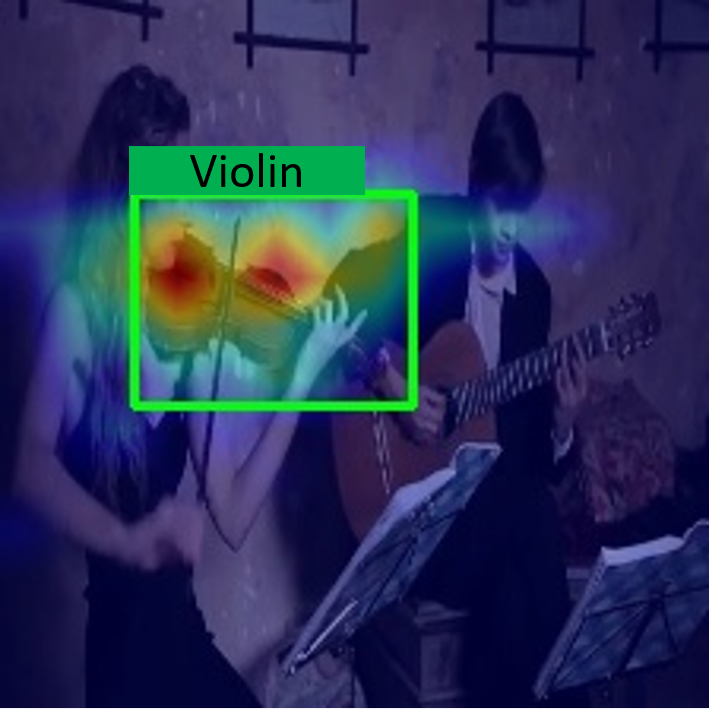}
    \hspace{1mm}
    \includegraphics[width=0.117\linewidth]{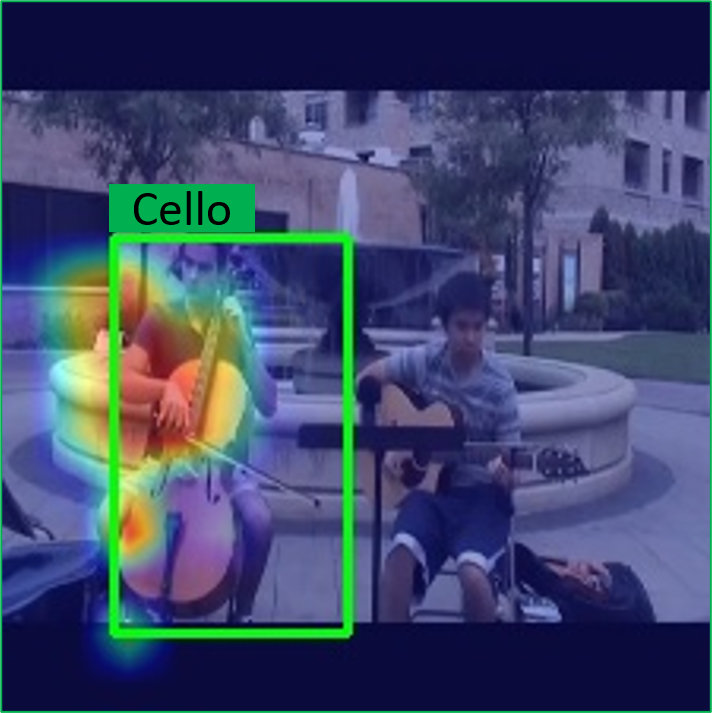}
    \includegraphics[width=0.117\linewidth]{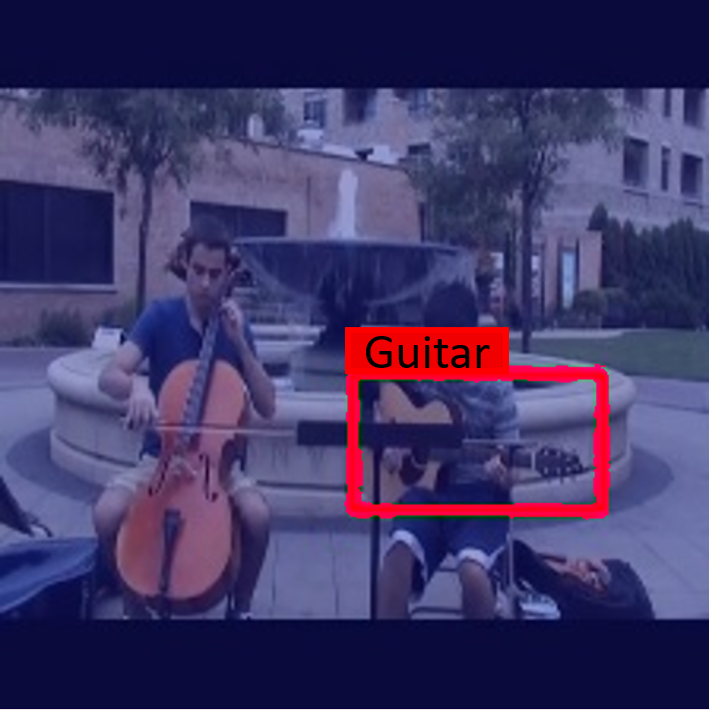}
    \hspace{1mm}
    \includegraphics[width=0.117\linewidth]{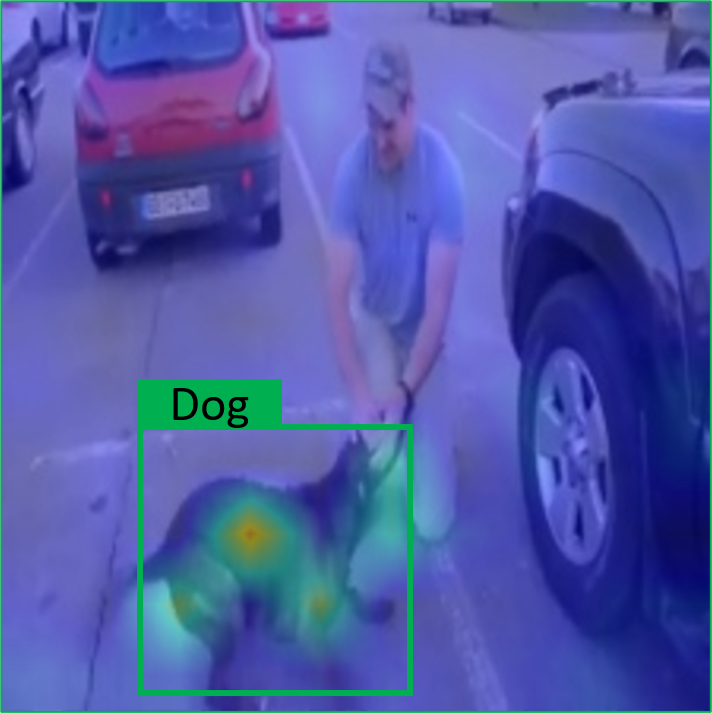}
    \includegraphics[width=0.117\linewidth]{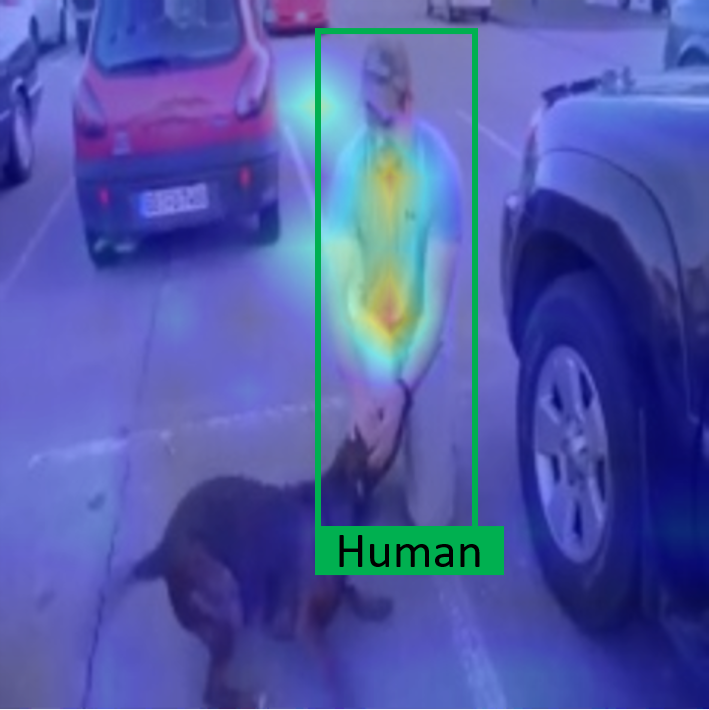}
    \includegraphics[width=0.117\linewidth]{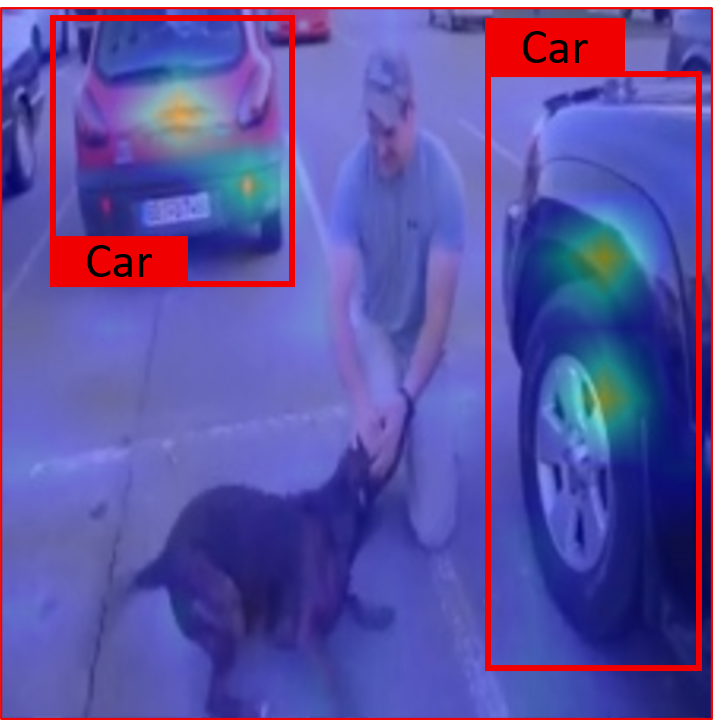}
    \includegraphics[width=0.117\linewidth]{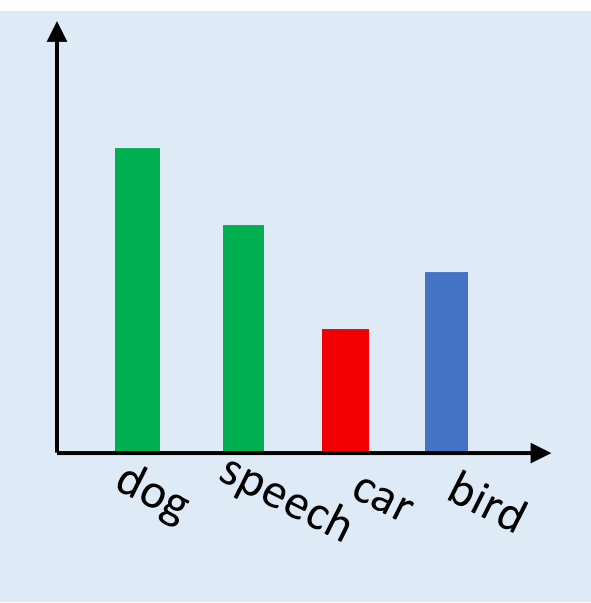}
    }\\
    \subfigure[DSOL~\cite{hu2020discriminative}]{
    \includegraphics[width=0.117\linewidth]{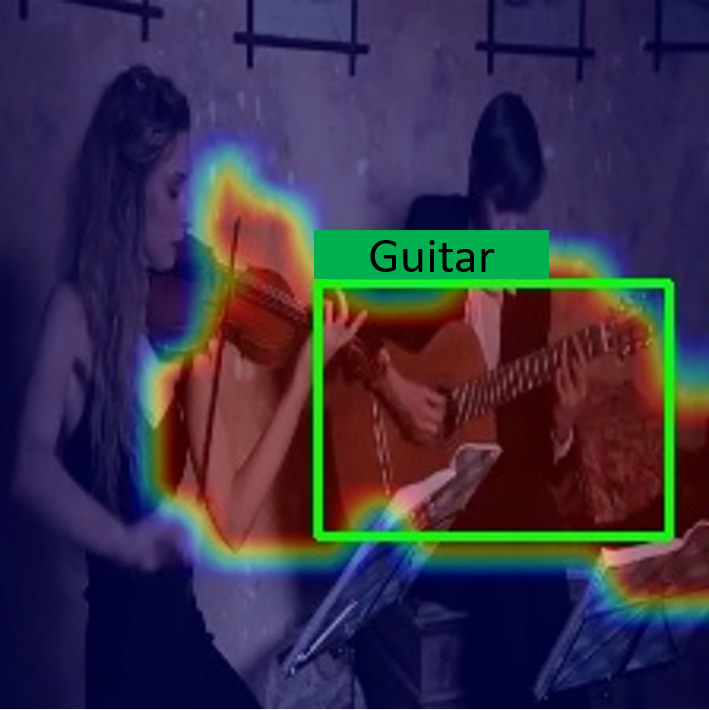}
    \includegraphics[width=0.117\linewidth]{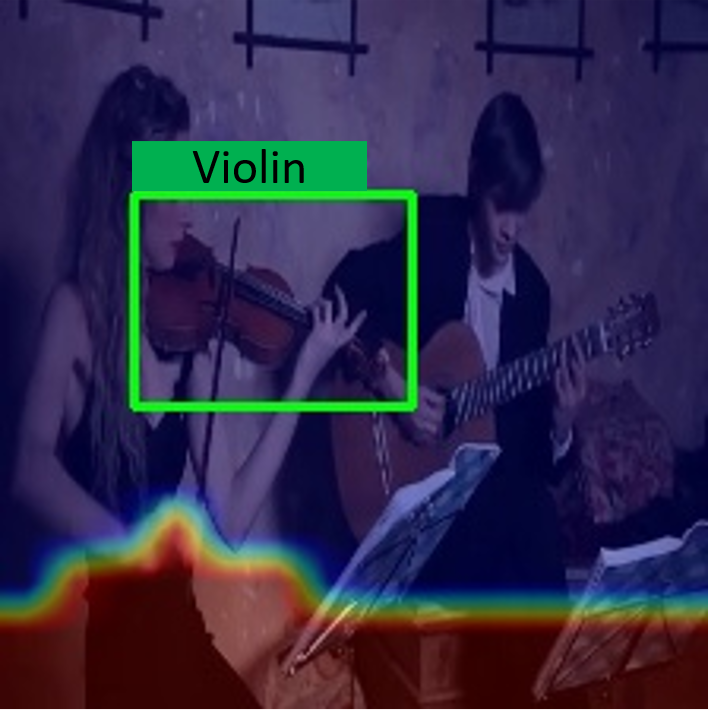}
    \hspace{1mm}
    \includegraphics[width=0.117\linewidth]{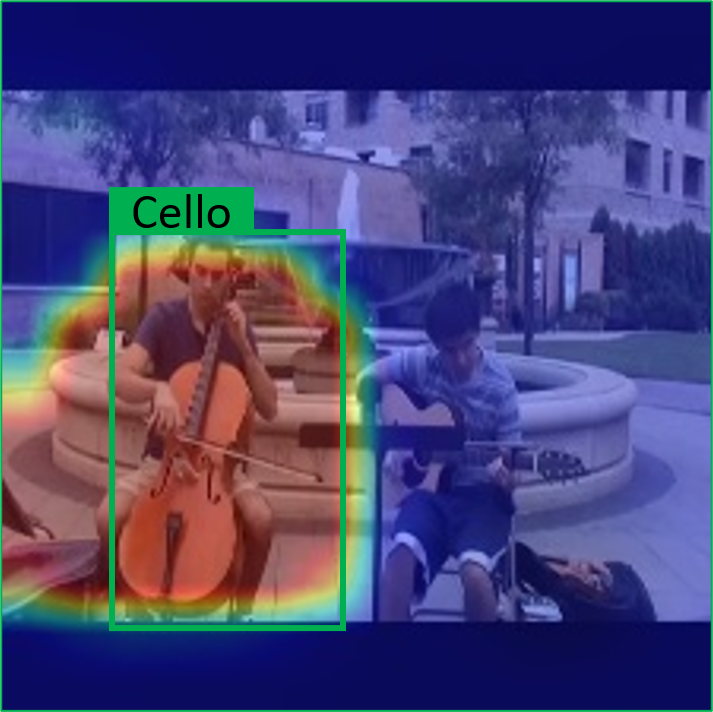}
    \includegraphics[width=0.117\linewidth]{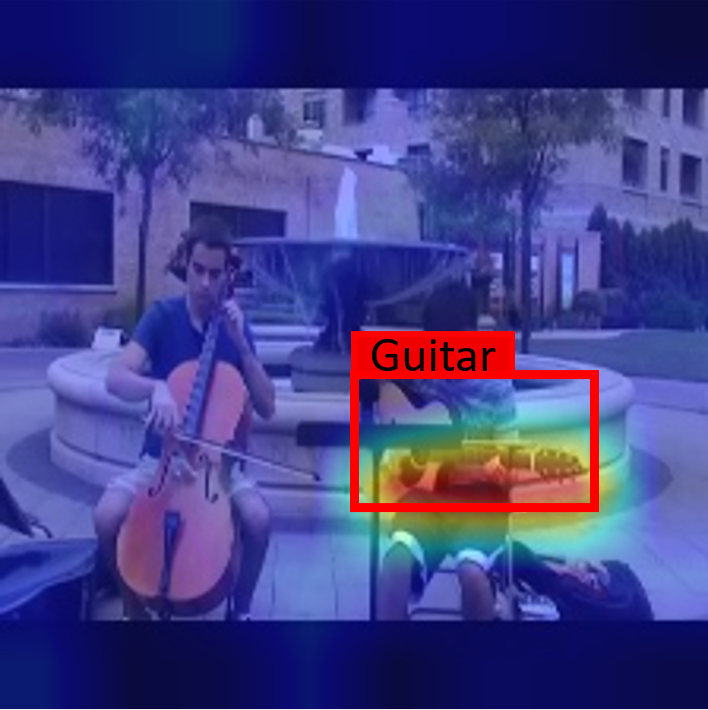}
    \hspace{1mm}
    \includegraphics[width=0.117\linewidth]{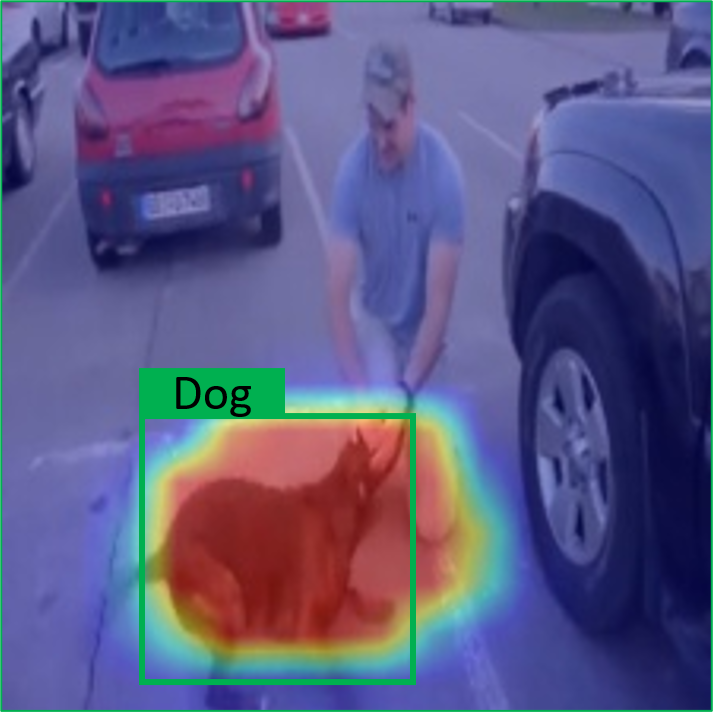}
    \includegraphics[width=0.117\linewidth]{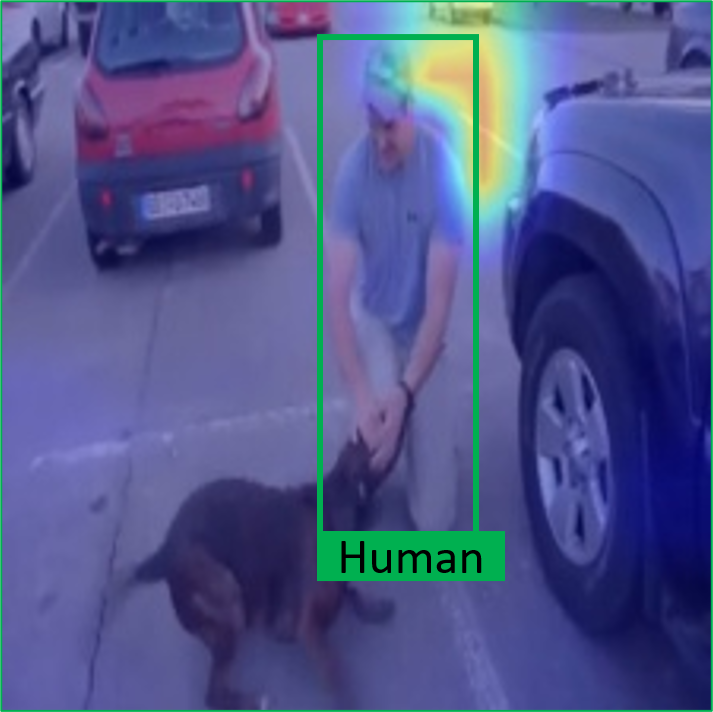}
    \includegraphics[width=0.117\linewidth]{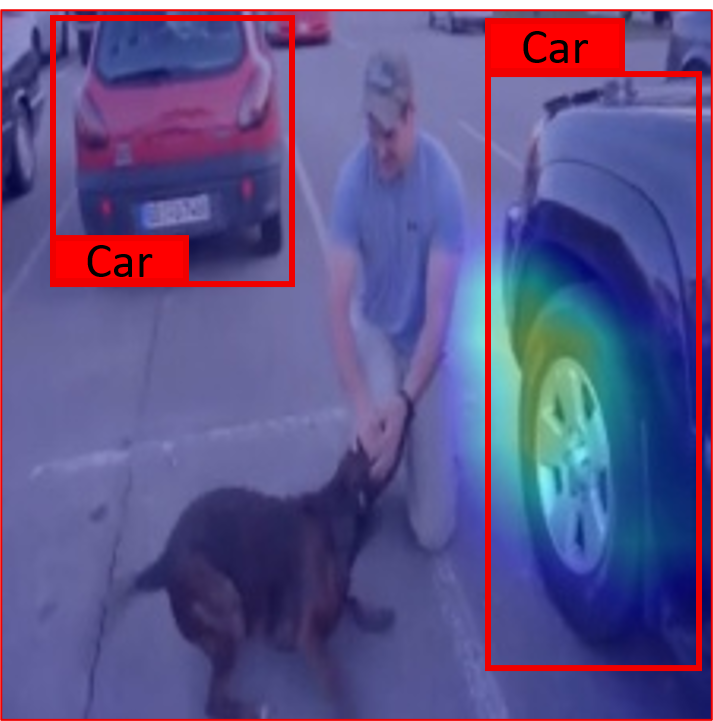}
    \includegraphics[width=0.117\linewidth]{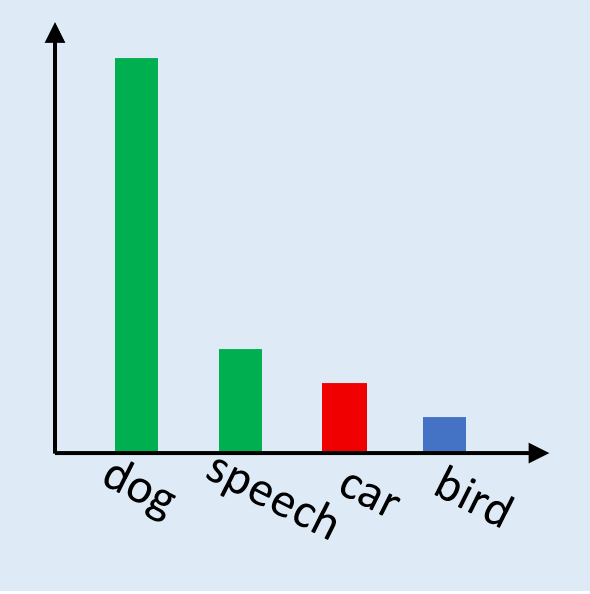}
    }\\
    \subfigure[Ours]{
    \includegraphics[width=0.117\linewidth]{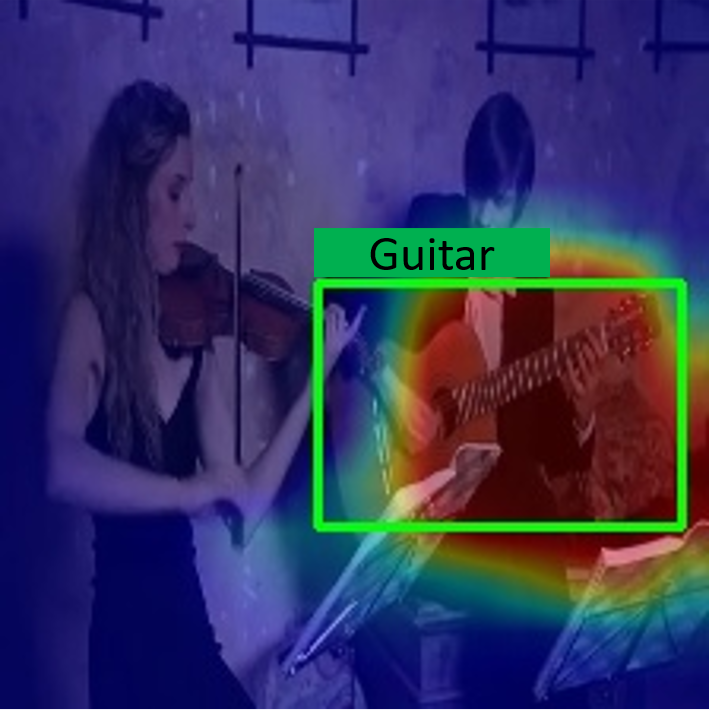}
    \includegraphics[width=0.117\linewidth]{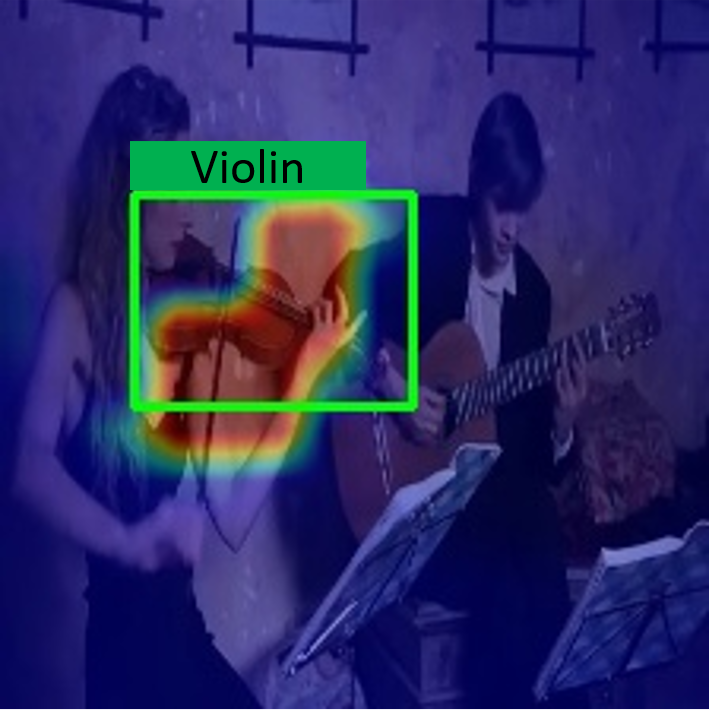}
    \hspace{1mm}
    \includegraphics[width=0.117\linewidth]{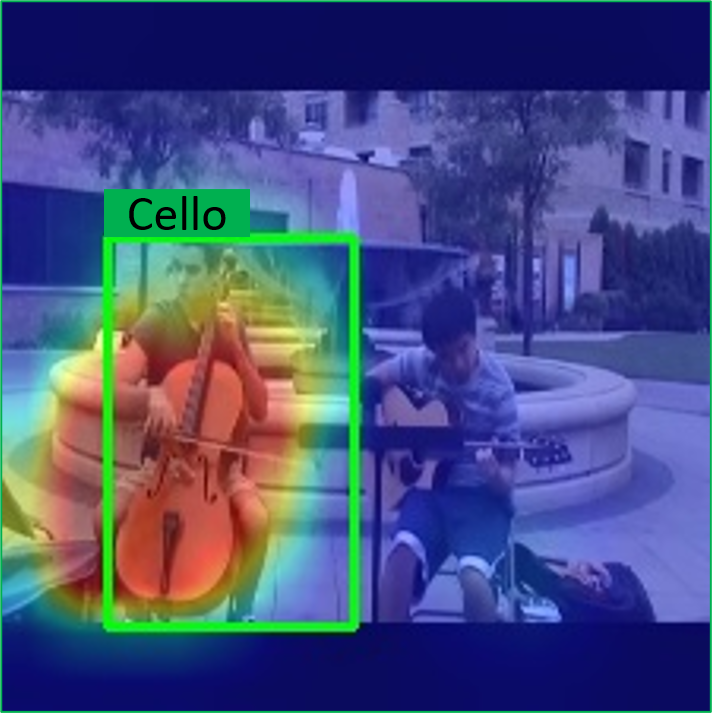}
    \includegraphics[width=0.117\linewidth]{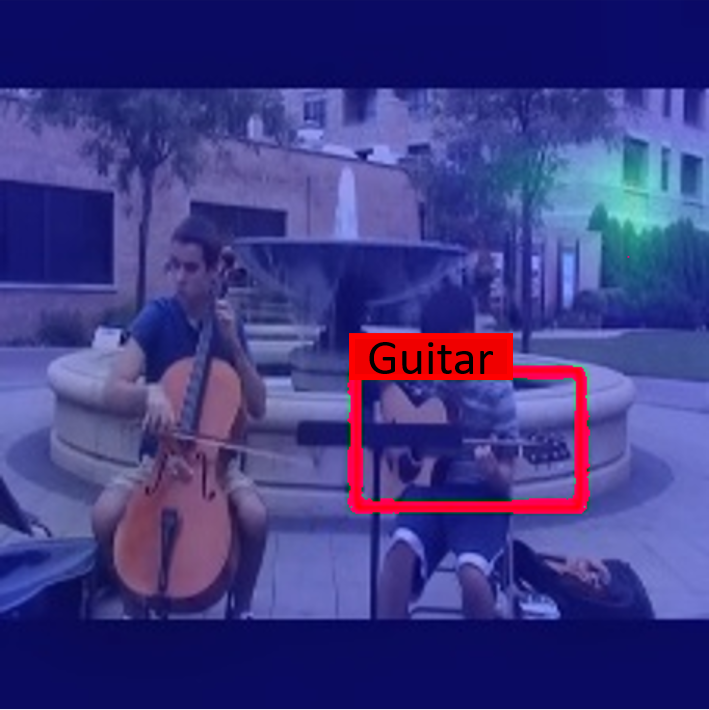}
    \hspace{1mm}
    \includegraphics[width=0.117\linewidth]{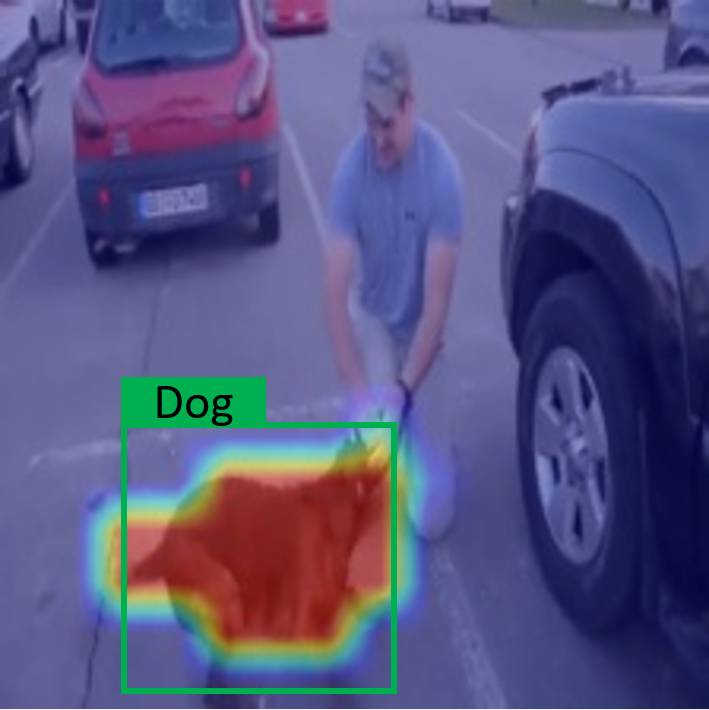}
    \includegraphics[width=0.117\linewidth]{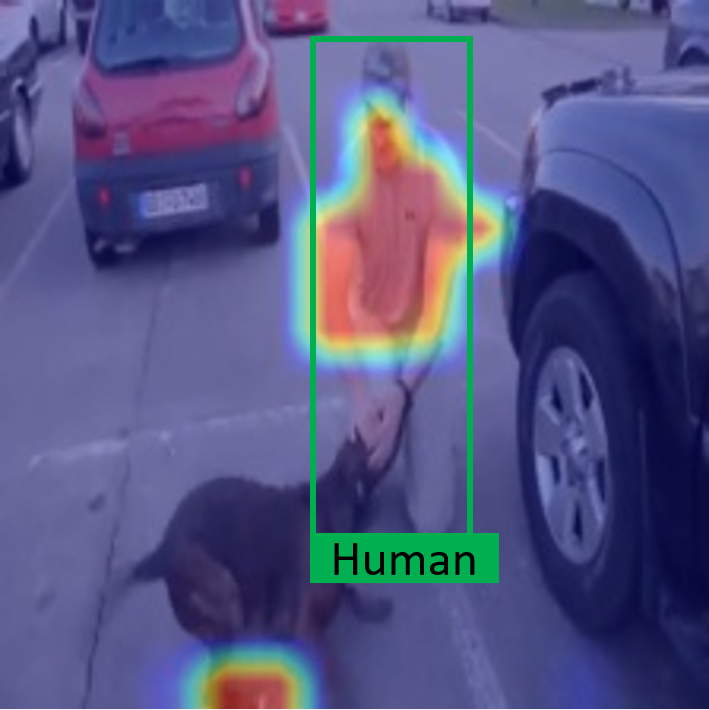}
    \includegraphics[width=0.117\linewidth]{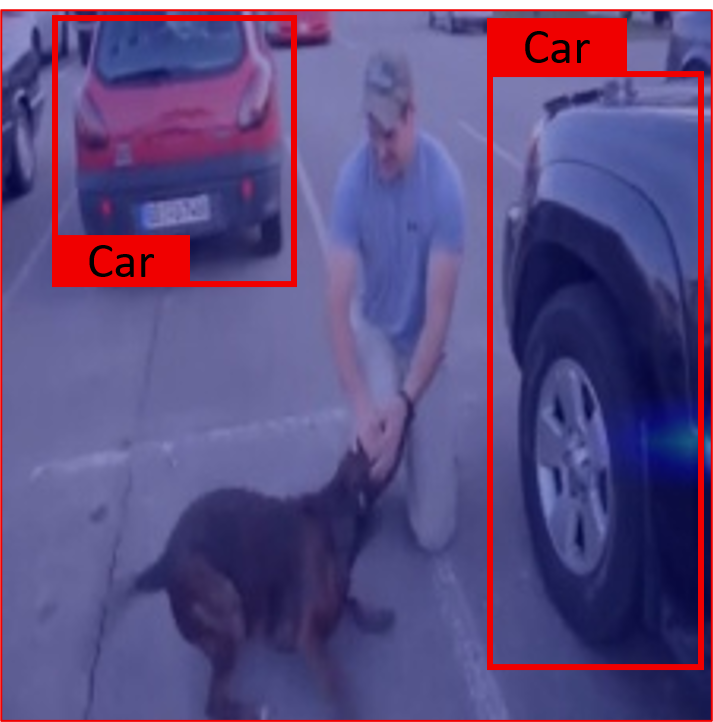}
    \includegraphics[width=0.117\linewidth]{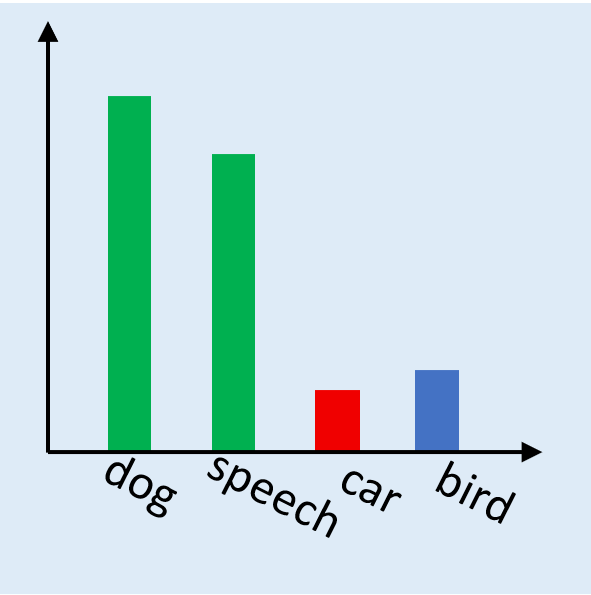}
    }
    \caption{\textbf{Comparison of sound localization and audio perception between different methods.} In class-specific localization maps, the green boxes are sounding objects, the red boxes indicate visible but silent objects. And the histogram represents perceived audio distribution, the green columns indicate in-the-scene sounds, the red column means silent category, and the blue column means off-screen sound that is expected to be suppressed. More qualitative results can be found in demo video.}
    \label{compare}
\end{figure*}

\subsection{Experimental Settings}

\noindent \textbf{Comparing Baselines.} We compare our model with the following methods: \textbf{Object-that-sound}~\cite{arandjelovic2018objects} which optimizes similarity between audio embedding and spatial visual features; \textbf{Sound-of-pixel}~\cite{zhao2018sound} that establishes channel-wise audio-visual correlation by training sound separation.
\cite{qian2020multiple} requires audio category prior to discriminate multiple sounds, which is different from our setting, so we do not make comparison. 
Particularly, we compare with the SOTA \textbf{DSOL}~\cite{hu2020discriminative}, which also learns object representation from single-sound data and then expands to complex cases.

Since we focus on visual sound localization without labels, where all classes are cluster-based pseudo ones. So we do not compare with other mixed audio perception methods.

\begin{table*}
    \centering
        \begin{tabular}{c|c|c|c|c|c|c|c|c}
            \hline
            Dataset & \multicolumn{4}{c|} {MUSIC} & \multicolumn{4}{c} {VGGSound} \\
            \hline
            Method & NMI & Prec. & Rec. & mAP & NMI & Prec. & Rec. & mAP\\
            \hline
            w/o \textbf{Audio-Instance-Identifier}  & 0.692 & 0.372 & 0.335 & 0.355 & 0.410 & 0.218 & 0.076 & 0.189 \\
            \textbf{Audio-Instance-Identifier} w/o curriculum  & 0.758 & 0.441 & 0.489 & 0.403 & 0.414 & 0.298 & 0.160 & 0.231\\
            \hline
            \textbf{Audio-Instance-Identifier} w/ curriculum & \textbf{0.809} & \textbf{0.461} & \textbf{0.715} & \textbf{0.433} & \textbf{0.436} & \textbf{0.346} & \textbf{0.232} & \textbf{0.283}\\
            \hline
        \end{tabular}
        \caption{Ablation study on Audio-Instance-Identifier with curriculum learning strategy in terms of NMI and multi-label audio classification. We empirically set the threshold as $0.5$ when calculating Prec. and Rec.}
        \label{nmi}
\end{table*}

\noindent \textbf{Evaluation Metrics.} For the multi-modal prototype extraction, we use \textit{Normalized Mutual Information} (NMI) between cluster assignments and category labels to measure the discrimination of audio-visual backbones. To validate model's ability in challenging mixed sound perception, following~\cite{vedaldi2020vggsound}, \emph{mean Average Precision} (mAP) is adopted to evaluate audio classification performance. Besides, we adopt \emph{Precision} (Prec.) and \emph{Recall} (Rec.) of multi-label classification for more detailed analysis.
\iffalse
\begin{align}
    \text{Precision} &= \frac{\sum^K_{t=1} (p^a(t)>\zeta)\mathcal{Y}^m(t)}{\sum_{t=1}^K p^a(t)>\zeta},
    \label{precision}
    \\
    \text{Recall} &= \frac{\sum_{t=1}^K (p^a(t)>\zeta)\mathcal{Y}^m(t)}{\sum_{t=1}^K \mathcal{Y}^m(t)},
    \label{recall}
\end{align}
where $p^a(t)$ is the probability of the $t$-th audio category, $\mathcal{Y}^m(t)$ is the pseudo-label for the $t$-th audio category and $\zeta$ indicates the threshold to determine whether the audio class exists in mixed sound.
\fi

For sound localization, we use \textit{Intersection over Union} (IoU) and \textit{Area Under Curve} (AUC)~\cite{senocak2018learning} for single-sound scene evaluation and adopt \textit{Class-aware IoU} (CIoU)~\cite{hu2020discriminative} for in-the-wild localization:
\begin{align}
    \text{CIoU} &= \frac{\sum_{t=1}^K \mathcal{Y}^m(t) {IoU}_t}{\sum_{t=1}^K \mathcal{Y}^m(t)},
    \label{ciou}
\end{align}
where $\mathcal{Y}^m(t)$ is the pseudo-label for the $t$-th category and ${IoU}_t$ is calculated based on the predicted sounding object area and annotated bounding box for the $t$-th class.

\noindent \textbf{Implementation Details.} We split videos in each dataset into non-overlapped 1-second clips to form audio-visual pairs. Concretely, we sample audio at 16kHz with window length of 160ms, hop length of 80ms and transform it into log-mel spectrograms with 64 frequency bins. For visual stream, we randomly sample a frame from each clip and resize it into 256$\times$256, then randomly/center crop into 224$\times$224 for training/evaluation. The model is trained by Adam optimizer with learning rate $10^{-4}$. For evaluation, we use Faster RCNN~\cite{ren2015faster} to detect bounding box as reference. We aggregate cluster assignments to category labels to class-specifically measure localization performance.

\subsection{Quantitative Analysis}

\noindent \textbf{Single Sound Localization.} We first analyze the quantitative result of single-sound localization. As shown in Table~\ref{tbl:syn}(a), our \textbf{IEr} framework outperforms existing methods on both MUSIC and VGGSound datasets. Considering that there still exists noise in single-sound data and not all audio-visual pairs are clean, such superiority indicates the \textbf{Audio-Instance-Identifier} trained on mixed audio improves the robustness and discrimination of extracted audio embedding.

\noindent \textbf{Discriminative Sound Localization in-the-Wild.} General in-the-wild scenarios are often accompanied by interference of unbalanced audio mixture, visible but silent objects and off-screen sound, which makes it intractable to localize sounding objects of each class without extra annotations. As shown in Table~\ref{tbl:syn}(b), our model outperforms other methods by a large margin on both realistic and synthetic scenes. Such significant improvements majorly benefit from two aspects: 1) The \textbf{Audio-Instance-Identifier} leverages the distinguishing-step with curriculum learning to obtain expanded class-specific mixed audio features, while~\cite{arandjelovic2018objects,zhao2019sound,hu2020discriminative} failed to handle mixed-sound perception well.
2) The \textbf{Cross-modal Referrer} eliminates the pervasive interference of visible but silent objects and off-screen noise in-the-wild, while previous methods ignored this problem and failed to achieve robust audio-visual matching.

\subsection{Qualitative Analysis}

Besides the quantitative analysis, we compare visualization results for class-specific sound localization and perceived in-the-scene audio distribution of different methods in Fig.~\ref{compare}. Note that Object-that-sound~\cite{arandjelovic2018objects} is only applicable to single-sound cases, we do not present their results. For in-the-wild scene in the right part of Fig.~\ref{compare}, Sound-of-pixel~\cite{zhao2018sound} cannot identify the exact location of objects and fails to filter out silent objects or suppress off-screen sound. This is because it requires clean data for the pretext task training and has difficulty adapting to noisy cases. The audio perception of DSOL~\cite{hu2020discriminative} is quite unbalanced and dominated by the dog barking, which makes it only manage to localize the dog while treating the talking person as silent. On the contrary, with the help of two key modules, our method achieves balanced audio perception as well as off-screen sound erasing. Besides, the proposed framework can precisely localize objects in a class-aware manner and filter out silent ones.

\subsection{Ablation Study}

\noindent\textbf{Audio-Instance-Identifier.} To quantify the efficacy of Audio-Instance-Identifier with curriculum learning strategy, we present the ablation study results in terms of NMI between cluster assignments and category labels as well as multi-label audio classification in Table~\ref{nmi}. It indicates that Audio-Instance-Identifier leads to improvement over all metrics. The significant gain in recall reveals that this module helps to avoid neglecting non-dominant sounds and balance audio perception. Besides, curriculum learning strategy further improves performance since it provides the model with effectively arranged audio samples to enhance the robustness to noise and the expressiveness of prototypes.

\begin{table}
    \centering
        \begin{tabular}{cc|cc|cc}
            \hline
            \multicolumn{2}{c|}{Dataset} & \multicolumn{2}{c|}{MUSIC-Un.} & \multicolumn{2}{c}{VGG-Un.} \\
            \hline
            Silent. & Off-Screen. & CIoU & AUC & CIoU & AUC\\
            \hline
            \XSolidBrush & \XSolidBrush  & 2.7 & 6.9 & 5.9 & 10.2 \\
            \CheckmarkBold & \XSolidBrush  & 4.6 & 9.5 & 7.2 & 14.1\\
            \XSolidBrush & \CheckmarkBold  & 5.1 & 10.8 & 7.8 & 13.9 \\
            \hline
            \CheckmarkBold & \CheckmarkBold  & \textbf{15.6} & \textbf{15.3} & \textbf{12.8} & \textbf{17.6}\\
            \hline
        \end{tabular}
        \caption{Ablation study on Silent Object Filter and Off-Screen Noise Filter for MUSIC-Un. and VGG-Un.}
        \label{tbl:misalign}   
\end{table}

\noindent\textbf{Cross-modal Referrer.} Table~\ref{tbl:misalign} shows the ablative results on two sub-modules of Cross-modal Referrer. The results demonstrate that it is necessary to erase interference in both modalities. That is, either off-screen sound or visible but silent objects introduce severe interference for audio-visual correspondence learning and corrupt sound localization.

\begin{table}
    \centering
        \begin{tabular}{cc|cc|cc}
            \hline
            \multicolumn{2}{c|}{Dataset} & \multicolumn{2}{c|}{MUSIC-Syn.} & \multicolumn{2}{c}{VGG-Un.} \\
            \hline
            AII. & CMR. & CIoU & AUC & CIoU & AUC\\
            \hline
      \XSolidBrush & \XSolidBrush  & 0.2 & 6.8 & 3.1 & 7.8 \\
      \CheckmarkBold & \XSolidBrush & 12.8 & 11.1 & 5.9 & 10.2 \\
      \XSolidBrush & \CheckmarkBold & 29.5 & 22.3 & 6.9 & 13.5 \\
            \hline
      \CheckmarkBold & \CheckmarkBold & \textbf{47.6} & \textbf{29.8} & \textbf{12.8} & \textbf{17.6}\\
            \hline
        \end{tabular}
        \caption{Ablative experiments on Audio-Instance-Identifier (AII.) and Cross-modal Referrer (CMR.).}
        \label{tbl:all}   
\end{table}

The overall ablation study of Audio-Instance-Identifier and Cross-modal Referrer is shown in Table~\ref{tbl:all}, verifying that both modules are crucial for in the wild sound localization.
\section{Conclusion}
In this work, we introduce a novel framework Interference Eraser (\textbf{IEr}) to enhance robust visual sound localization for in-the-wild scenes. We identify the two strengths of our method: 1) Our proposed Audio-Instance-Identifier learns the distinguishing-step to achieve volume agnostic mixed sound perception, which erases the interference of uneven sound mixtures. 2) We propose the Cross-modal Referrer to eliminate the interference of visible but silent objects and audible but off-screen sounds. Extensive experiments demonstrate the superiority of our proposed method for sound localization, particularly for in-the-wild scenarios.

\noindent\textbf{Acknowledgment.} This study is supported by National Natural Science Foundation of China (No. U21B2013), NTU NAP, MOE AcRF Tier 1 (2021-T1-001-088), and under the RIE2020 Industry Alignment Fund – Industry Collaboration Projects (IAF-ICP) Funding Initiative, as well as cash and in-kind contribution from the industry partner(s).

% \bibliography{aaai22}

{\small
\bibliography{ref}

\begin{thebibliography}{39}
\providecommand{\natexlab}[1]{#1}

\bibitem[{Adavanne et~al.(2018)Adavanne, Politis, Nikunen, and
  Virtanen}]{adavanne2018sound}
Adavanne, S.; Politis, A.; Nikunen, J.; and Virtanen, T. 2018.
\newblock Sound event localization and detection of overlapping sources using
  convolutional recurrent neural networks.
\newblock \emph{IEEE Journal of Selected Topics in Signal Processing}, 13(1):
  34--48.

\bibitem[{Afouras et~al.(2020)Afouras, Owens, Chung, and
  Zisserman}]{afouras2020self}
Afouras, T.; Owens, A.; Chung, J.~S.; and Zisserman, A. 2020.
\newblock Self-supervised learning of audio-visual objects from video.
\newblock In \emph{Computer Vision--ECCV 2020: 16th European Conference,
  Glasgow, UK, August 23--28, 2020, Proceedings, Part XVIII 16}, 208--224.
  Springer.

\bibitem[{Alwassel et~al.(2019)Alwassel, Mahajan, Torresani, Ghanem, and
  Tran}]{alwassel2019self}
Alwassel, H.; Mahajan, D.; Torresani, L.; Ghanem, B.; and Tran, D. 2019.
\newblock Self-supervised learning by cross-modal audio-video clustering.
\newblock \emph{arXiv preprint arXiv:1911.12667}.

\bibitem[{Arandjelovic and Zisserman(2017)}]{arandjelovic2017look}
Arandjelovic, R.; and Zisserman, A. 2017.
\newblock Look, listen and learn.
\newblock In \emph{Proceedings of the IEEE International Conference on Computer
  Vision}, 609--617.

\bibitem[{Arandjelovic and Zisserman(2018)}]{arandjelovic2018objects}
Arandjelovic, R.; and Zisserman, A. 2018.
\newblock Objects that sound.
\newblock In \emph{Proceedings of the European Conference on Computer Vision
  (ECCV)}, 435--451.

\bibitem[{Aytar, Vondrick, and Torralba(2016)}]{aytar2016soundnet}
Aytar, Y.; Vondrick, C.; and Torralba, A. 2016.
\newblock Soundnet: Learning sound representations from unlabeled video.
\newblock In \emph{Advances in neural information processing systems},
  892--900.

\bibitem[{Caron et~al.(2018)Caron, Bojanowski, Joulin, and
  Douze}]{caron2018deep}
Caron, M.; Bojanowski, P.; Joulin, A.; and Douze, M. 2018.
\newblock Deep clustering for unsupervised learning of visual features.
\newblock In \emph{Proceedings of the European Conference on Computer Vision
  (ECCV)}, 132--149.

\bibitem[{Chen et~al.(2021)Chen, Xie, Afouras, Nagrani, Vedaldi, and
  Zisserman}]{chen2021localizing}
Chen, H.; Xie, W.; Afouras, T.; Nagrani, A.; Vedaldi, A.; and Zisserman, A.
  2021.
\newblock Localizing Visual Sounds the Hard Way.
\newblock In \emph{Proceedings of the IEEE/CVF Conference on Computer Vision
  and Pattern Recognition}, 16867--16876.

\bibitem[{Gan et~al.(2020)Gan, Huang, Zhao, Tenenbaum, and
  Torralba}]{gan2020music}
Gan, C.; Huang, D.; Zhao, H.; Tenenbaum, J.~B.; and Torralba, A. 2020.
\newblock Music Gesture for Visual Sound Separation.
\newblock In \emph{Proceedings of the IEEE/CVF Conference on Computer Vision
  and Pattern Recognition}, 10478--10487.

\bibitem[{Gao and Grauman(2019)}]{gao2019co}
Gao, R.; and Grauman, K. 2019.
\newblock Co-separating sounds of visual objects.
\newblock In \emph{Proceedings of the IEEE International Conference on Computer
  Vision}, 3879--3888.

\bibitem[{He et~al.(2016)He, Zhang, Ren, and Sun}]{he2016deep}
He, K.; Zhang, X.; Ren, S.; and Sun, J. 2016.
\newblock Deep residual learning for image recognition.
\newblock In \emph{Proceedings of the IEEE conference on computer vision and
  pattern recognition}, 770--778.

\bibitem[{Hinton, Vinyals, and Dean(2015)}]{44873}
Hinton, G.; Vinyals, O.; and Dean, J. 2015.
\newblock Distilling the Knowledge in a Neural Network.
\newblock In \emph{NIPS Deep Learning and Representation Learning Workshop}.

\bibitem[{Hu et~al.(2020)Hu, Qian, Jiang, Tan, Wen, Ding, Lin, and
  Dou}]{hu2020discriminative}
Hu, D.; Qian, R.; Jiang, M.; Tan, X.; Wen, S.; Ding, E.; Lin, W.; and Dou, D.
  2020.
\newblock Discriminative Sounding Objects Localization via Self-supervised
  Audiovisual Matching.
\newblock \emph{arXiv preprint arXiv:2010.05466}.

\bibitem[{Korbar, Tran, and Torresani(2018)}]{korbar2018cooperative}
Korbar, B.; Tran, D.; and Torresani, L. 2018.
\newblock Cooperative learning of audio and video models from self-supervised
  synchronization.
\newblock In \emph{Advances in Neural Information Processing Systems},
  7763--7774.

\bibitem[{Lin et~al.(2017)Lin, Doll{\'a}r, Girshick, He, Hariharan, and
  Belongie}]{lin2017feature}
Lin, T.-Y.; Doll{\'a}r, P.; Girshick, R.; He, K.; Hariharan, B.; and Belongie,
  S. 2017.
\newblock Feature pyramid networks for object detection.
\newblock In \emph{Proceedings of the IEEE conference on computer vision and
  pattern recognition}, 2117--2125.

\bibitem[{Lin et~al.(2021)Lin, Tseng, Lee, Lin, and Yang}]{lin2021unsupervised}
Lin, Y.-B.; Tseng, H.-Y.; Lee, H.-Y.; Lin, Y.-Y.; and Yang, M.-H. 2021.
\newblock Unsupervised Sound Localization via Iterative Contrastive Learning.
\newblock \emph{arXiv preprint arXiv:2104.00315}.

\bibitem[{Owens and Efros(2018)}]{owens2018audio}
Owens, A.; and Efros, A.~A. 2018.
\newblock Audio-visual scene analysis with self-supervised multisensory
  features.
\newblock In \emph{Proceedings of the European Conference on Computer Vision
  (ECCV)}, 631--648.

\bibitem[{Patrick et~al.(2020)Patrick, Asano, Fong, Henriques, Zweig, and
  Vedaldi}]{patrick2020multi}
Patrick, M.; Asano, Y.~M.; Fong, R.; Henriques, J.~F.; Zweig, G.; and Vedaldi,
  A. 2020.
\newblock Multi-modal self-supervision from generalized data transformations.
\newblock \emph{arXiv preprint arXiv:2003.04298}.

\bibitem[{Phan et~al.(2017)Phan, Koch, Katzberg, Maass, Mazur, McLoughlin, and
  Mertins}]{phan2017makes}
Phan, H.; Koch, P.; Katzberg, F.; Maass, M.; Mazur, R.; McLoughlin, I.; and
  Mertins, A. 2017.
\newblock What makes audio event detection harder than classification?
\newblock In \emph{2017 25th European signal processing conference (EUSIPCO)},
  2739--2743. IEEE.

\bibitem[{Proulx et~al.(2014)Proulx, Brown, Pasqualotto, and
  Meijer}]{proulx2014multisensory}
Proulx, M.~J.; Brown, D.~J.; Pasqualotto, A.; and Meijer, P. 2014.
\newblock Multisensory perceptual learning and sensory substitution.
\newblock \emph{Neuroscience \& Biobehavioral Reviews}, 41: 16--25.

\bibitem[{Pu et~al.(2017)Pu, Panagakis, Petridis, and Pantic}]{pu2017audio}
Pu, J.; Panagakis, Y.; Petridis, S.; and Pantic, M. 2017.
\newblock Audio-visual object localization and separation using low-rank and
  sparsity.
\newblock In \emph{2017 IEEE International Conference on Acoustics, Speech and
  Signal Processing (ICASSP)}, 2901--2905. IEEE.

\bibitem[{Qian et~al.(2020)Qian, Hu, Dinkel, Wu, Xu, and
  Lin}]{qian2020multiple}
Qian, R.; Hu, D.; Dinkel, H.; Wu, M.; Xu, N.; and Lin, W. 2020.
\newblock Multiple Sound Sources Localization from Coarse to Fine.
\newblock \emph{arXiv preprint arXiv:2007.06355}.

\bibitem[{Ren et~al.(2015)Ren, He, Girshick, and Sun}]{ren2015faster}
Ren, S.; He, K.; Girshick, R.; and Sun, J. 2015.
\newblock Faster r-cnn: Towards real-time object detection with region proposal
  networks.
\newblock In \emph{Advances in neural information processing systems}, 91--99.

\bibitem[{Senocak et~al.(2018)Senocak, Oh, Kim, Yang, and
  So~Kweon}]{senocak2018learning}
Senocak, A.; Oh, T.-H.; Kim, J.; Yang, M.-H.; and So~Kweon, I. 2018.
\newblock Learning to localize sound source in visual scenes.
\newblock In \emph{Proceedings of the IEEE Conference on Computer Vision and
  Pattern Recognition}, 4358--4366.

\bibitem[{Stein and Meredith(1993)}]{stein1993merging}
Stein, B.~E.; and Meredith, M.~A. 1993.
\newblock \emph{The merging of the senses.}
\newblock The MIT Press.

\bibitem[{Tian, Hu, and Xu(2021)}]{tian2021cyclic}
Tian, Y.; Hu, D.; and Xu, C. 2021.
\newblock Cyclic Co-Learning of Sounding Object Visual Grounding and Sound
  Separation.
\newblock In \emph{Proceedings of the IEEE/CVF Conference on Computer Vision
  and Pattern Recognition}, 2745--2754.

\bibitem[{Tian et~al.(2018)Tian, Shi, Li, Duan, and Xu}]{tian2018audio}
Tian, Y.; Shi, J.; Li, B.; Duan, Z.; and Xu, C. 2018.
\newblock Audio-visual event localization in unconstrained videos.
\newblock In \emph{Proceedings of the European Conference on Computer Vision
  (ECCV)}, 247--263.

\bibitem[{Tian and Xu(2021)}]{tian2021can}
Tian, Y.; and Xu, C. 2021.
\newblock Can audio-visual integration strengthen robustness under multimodal
  attacks?
\newblock In \emph{Proceedings of the IEEE/CVF Conference on Computer Vision
  and Pattern Recognition}, 5601--5611.

\bibitem[{Tzinis et~al.(2021)Tzinis, Wisdom, Remez, and
  Hershey}]{tzinis2021improving}
Tzinis, E.; Wisdom, S.; Remez, T.; and Hershey, J.~R. 2021.
\newblock Improving On-Screen Sound Separation for Open Domain Videos with
  Audio-Visual Self-attention.
\newblock \emph{arXiv preprint arXiv:2106.09669}.

\bibitem[{Vedaldi et~al.(2020)Vedaldi, Zisserman, Chen, and
  Xie}]{vedaldi2020vggsound}
Vedaldi, A.; Zisserman, A.; Chen, H.; and Xie, W. 2020.
\newblock Vggsound: a large-scale audio-visual dataset.
\newblock In \emph{Proceedings of the International Conference on Acoustics,
  Speech, and Signal Processing}. IEEE.

\bibitem[{Xu et~al.(2021)Xu, Zhou, Liu, Dai, Wang, and Lin}]{xu2021visually}
Xu, X.; Zhou, H.; Liu, Z.; Dai, B.; Wang, X.; and Lin, D. 2021.
\newblock Visually Informed Binaural Audio Generation without Binaural Audios.
\newblock In \emph{Proceedings of the IEEE conference on computer vision and
  pattern recognition (CVPR)}.

\bibitem[{Yosinski et~al.(2014)Yosinski, Clune, Bengio, and
  Lipson}]{yosinski2014transferable}
Yosinski, J.; Clune, J.; Bengio, Y.; and Lipson, H. 2014.
\newblock How transferable are features in deep neural networks?
\newblock In \emph{Advances in neural information processing systems},
  3320--3328.

\bibitem[{Zhao et~al.(2019)Zhao, Gan, Ma, and Torralba}]{zhao2019sound}
Zhao, H.; Gan, C.; Ma, W.-C.; and Torralba, A. 2019.
\newblock The sound of motions.
\newblock In \emph{Proceedings of the IEEE International Conference on Computer
  Vision}, 1735--1744.

\bibitem[{Zhao et~al.(2018)Zhao, Gan, Rouditchenko, Vondrick, McDermott, and
  Torralba}]{zhao2018sound}
Zhao, H.; Gan, C.; Rouditchenko, A.; Vondrick, C.; McDermott, J.; and Torralba,
  A. 2018.
\newblock The sound of pixels.
\newblock In \emph{Proceedings of the European conference on computer vision
  (ECCV)}, 570--586.

\bibitem[{Zhou et~al.(2019{\natexlab{a}})Zhou, Liu, Liu, Luo, and
  Wang}]{zhou2019talking}
Zhou, H.; Liu, Y.; Liu, Z.; Luo, P.; and Wang, X. 2019{\natexlab{a}}.
\newblock Talking Face Generation by Adversarially Disentangled Audio-Visual
  Representation.
\newblock In \emph{AAAI Conference on Artificial Intelligence (AAAI)}.

\bibitem[{Zhou et~al.(2019{\natexlab{b}})Zhou, Liu, Xu, Luo, and
  Wang}]{Zhou_2019_ICCV}
Zhou, H.; Liu, Z.; Xu, X.; Luo, P.; and Wang, X. 2019{\natexlab{b}}.
\newblock Vision-Infused Deep Audio Inpainting.
\newblock In \emph{The IEEE International Conference on Computer Vision
  (ICCV)}.

\bibitem[{Zhou et~al.(2021)Zhou, Sun, Wu, Loy, Wang, and Liu}]{zhou2021pose}
Zhou, H.; Sun, Y.; Wu, W.; Loy, C.~C.; Wang, X.; and Liu, Z. 2021.
\newblock Pose-controllable talking face generation by implicitly modularized
  audio-visual representation.
\newblock In \emph{Proceedings of the IEEE/CVF Conference on Computer Vision
  and Pattern Recognition}, 4176--4186.

\bibitem[{Zhou et~al.(2020)Zhou, Xu, Lin, Wang, and Liu}]{zhou2020sep}
Zhou, H.; Xu, X.; Lin, D.; Wang, X.; and Liu, Z. 2020.
\newblock Sep-Stereo: Visually Guided Stereophonic Audio Generation by
  Associating Source Separation.
\newblock In \emph{Proceedings of the European Conference on Computer Vision
  (ECCV)}.

\bibitem[{Zou et~al.(2020)Zou, Zhang, Moura, Yu, and Tian}]{zou2020revisiting}
Zou, Y.; Zhang, S.; Moura, J.~M.; Yu, J.; and Tian, Y. 2020.
\newblock Revisiting Mid-Level Patterns for Distant-Domain Few-Shot
  Recognition.
\newblock \emph{arXiv preprint arXiv:2008.03128}.

\end{thebibliography}
}

\clearpage
\appendix
\twocolumn[{%
\begin{minipage}{\textwidth}
   \null
   \vspace*{0.375in}
   \begin{center}
      {\Large \bf Supplemental Document:\\Visual Sound Localization in the Wild by Cross-Modal Interference Erasing \par}
   \end{center}
   \vspace*{0.375in}
\end{minipage}
\vspace{-6mm}}]

\section{Preliminary Notations}
To make the narration clearer, we first review the notations used in the paper. 

The training videos are denoted as $\mathcal{X} = \{(a_i, v_i) | i = 1, 2, ..., N\}$, where $N$ is the number of videos and $(a_i, v_i)$ is the $i$-th audio-visual pair. $\mathcal{X}$ can be divided into two disjoint subsets: $\mathcal{X}^s = \{(a_i^s, v_i^s) | i = 1, 2, ..., N^s\}$ consisting of $N^s$ single source videos and $\mathcal{X}^u = \{(a_i^u, v_i^u) | i = 1, 2, ..., N^u\}$ consisting of $N^u$ unconstrained in-the-wild videos.

\noindent\textbf{Audio-Visual Prototype Extraction.} We employ the variants of ResNet-18~\cite{he2016deep} as audio and visual backbones to extract visual feature map $f^v\in \mathbb{R}^{H\times W \times C}
$ and the audio feature $f^a \in \mathbb{R}^{C}$ from $\mathcal{X}^s$. The cosine similarity can be computed between each positional feature $f^{v_{(x,y)}}$ on $f^v$ and $f^a$ to present an audio-visual association map (localization map) $l(f^v, f^a) \in \mathbb{R}^{H\times W}$ with the audio-visual correspondence loss function $\mathcal{L}_{av}$. Then, we utilize the deep clustering on visual feature and extract the centroid of each class as \emph{prototype}. Based on one-to-one audio-visual pair correlation, the visual prototypes $\mathcal{P}^v \in \mathbb{R}^{K\times C}$ and audio prototypes $\mathcal{P}^a \in \mathbb{R}^{K\times C}$ can be established.

After the audio-visual prototype extraction process on single-sound training dataset $\mathcal{X}^s$, we can train the classification task on both audio and visual modalities. Recall that the deep visual cluster provides $K$ clusters. Thus for each audio-visual data pair $(a^{s_k}_i, v^{s_k}_i)$ in $\mathcal{X}^s$, a pseudo-class $k$ can be assigned. The pseudo label for audio $a^{s_k}_i$ can be represented as $\mathcal{Y}_i \in \mathbb{R}^{K}$ whose $k$-th element $\mathcal{Y}^k_i = 1$ and others $\mathcal{Y}^n_i = 0$ when $n \neq k$. By training solely on $\mathcal{X}^s$,  we can write the objective for this single-source audio clip as:

\begin{align}
    \mathcal{L}^s_p &= \frac{1}{K}\sum^{K}_{n=1} \mathcal{L}_{bce}(sim({\mathcal{P}^a_n}, f^a_i), \mathcal{Y}^n_i).
    \label{sup-eq:single}
\end{align}
where $sim(f_1, f_2)$ is the cosine similarity function defined as $sim(f_1, f_2) = \frac{\bm{f}_1\cdot \bm{f}_2}{|\bm{f}_1||\bm{f}_2|}$.

\noindent\textbf{Audio-Instance-Identifier.} We firstly train on single-sound audio data with loss function $\mathcal{L}_p^s$, and then learn the distinguishing-step $\Delta^n(a)$ with loss function $\mathcal{L}_p^m$ through curriculum learning strategy. 

\noindent\textbf{Cross-modal Referrer.} The class-specific visual localization map for each category $k$ is $L^v_k \in \mathbb{R}^{H \times W}$. With the predicted distinguishing-step $\boldsymbol{\Delta}(a)$, we can get the expanded class-specific features ${F}^a_k$ and calculate the class-aware audio-visual localization map $L^{av}$. The binary object mask $m$ regularizes the audio-appearing distribution $\text{p}^a$ to the audio-guided distribution $\text{p}^{av}$ and matches with the visual-guided distribution $\text{p}^{va}$ through cross-modal distillation loss $\mathcal{L}_u$. 

\section{Details of the Distinguishing-step}

In this section, we present the details of the distinguishing-step in Audio-Instance-Identifier. As mentioned in the paper, we employ weighted combination of linearly transformed mid-level features to predict the distinguishing-step. The \textbf{intuition} is: The input spectrogram of audio data or low-level audio feature are still not well extracted, while the high-level audio feature is compact in both time and frequency dimension (the dimension of feature embedding is decreasing, so it is becoming more and more compact), making features of different audio instances highly-overlapped. Therefore, directly using high-level audio feature is difficult to achieve balanced perception. In contrary, mid-level audio feature has advantages in distinguishing-step prediction in two aspects: 1) The feature embedding is better extracted than low-level audio feature, which makes it more expressive; 2) The feature embedding is less affected by the dominant sound than the compact highly-overlapped high-level feature, which makes it easier to capture more fine-grained information as also mentioned in~\cite{yosinski2014transferable,lin2017feature,zou2020revisiting}.

Specifically, we take the two-mix (the mixed audio sample consists of two different kinds of single sound signals) as an example: Given a mixed-sound audio sample $a_{ij}$, we use the audio backbone to extract deep representation $f^a_{ij}$ as well as mid-level feature $f_m(a_{ij})\in\mathbb{R}^{C_m}$.

To transform the mid-level feature into a joint embedding space, we define a set of linear transformation matrices $W_n\in\mathbb{R}^{C\times C_m}$ as well as biases $b_n\in\mathbb{R}^C$, where $n$ denotes category index. And the transformation is formulated as
\begin{align}
    \Delta^n(a_{ij}) = W_n f_m(a_{ij}) + b_n,
\end{align}
where $\Delta^n(a_{ij})$ denotes the linearly transformed mid-level feature of $n$-th category.
In this way, we obtain the predicted expanded class-specific representation of each category as $f^a_{ij}+\Delta^n(a_{ij})$.

In the training phase, we use manually mixed audio sample $a_{ij}$ with its pseudo category $\mathcal{Y}_{ij}$ and learned audio prototypes $\mathcal{P}^a$ to calculate the target distinguishing-step $\Delta(a_{ij})$. We minimize the multi-class classification loss in Eq.~\ref{sup-residual} for optimization:
\begin{align}
    \mathcal{L}^m_p &= \frac{1}{K}\sum^{K}_{n=1} \mathcal{L}_{bce}(sim({\mathcal{P}^a_n}, f^a_{ij} + \Delta^n(a)), \mathcal{Y}^n_{ij}).
    \label{sup-residual}
\end{align}
where $sim(f_1, f_2)$ is the cosine similarity function defined as $sim(f_1, f_2) = \frac{\bm{f}_1\cdot \bm{f}_2}{|\bm{f}_1||\bm{f}_2|}$.

In the inference phase, we adopt the learned parameters to predict the distinguishing-step on unconstrained audio data $a^u$ and generate $\Delta(a_{ij})$, which is then added to $f_{ij}^a$ to generate expanded class-specific features ${F}^a_k = f^a + {\Delta}^k(a)$.

\begin{table*}
	\centering
	\renewcommand{\arraystretch}{1.1}
  \begin{tabular}{l|cc|cc|cc|cc}
    \toprule[1.2pt]
    \makecell[c]{Dataset} & 
    \multicolumn{2}{c|}{MUSIC-Syn.} & \multicolumn{2}{c|}{MUSIC-Duet} & \multicolumn{2}{c|}{MUSIC-Un.} & \multicolumn{2}{c}{VGG-Un.} \\\hline
    \makecell[c]{Methods} & CIoU & AUC & CIoU & AUC & CIoU & AUC & CIoU & AUC\\
    \hline 
     \makecell[c]{Eq.~\ref{sup-1}} & 34.7 & 24.8 & 32.9  & 22.7 & 11.6 & 12.3 & 8.9 & 14.7 \\    
     \makecell[c]{Eq.~\ref{sup-2}} & 32.6  & 23.8  & 34.8 & 23.9 & 12.1 & 10.9 & 9.2 & 13.8\\
     \makecell[c]{Eq.~\ref{sup-3}} & 26.9  & 20.1 & 27.3  & 20.5 & 4.5 & 8.1 & 6.7 & 10.4 \\
     \makecell[c]{Eq.~\ref{sup-4}} & 38.1  & 26.2 & 41.4 & 26.8 & 11.4 & 11.1 & 9.1 & 15.3 \\
     \hline
     \makecell[c]{Eq.~\ref{sup-5_1} + Eq.~\ref{sup-5_2}} & \textbf{47.6}  & \textbf{29.8} & \textbf{52.9}   & \textbf{33.8} & \textbf{15.6} & \textbf{15.3} & \textbf{12.8} & \textbf{17.6} \\
     \bottomrule[1.2pt]
  \end{tabular}
%   \vspace{-1mm}  
  \caption{\textbf{Experiments on five alternative thresholds (methods) for binarized mask in MUSIC-Syn., MUSIC-Duet, MUSIC-Un. and VGG-Un. Dataset.} We compare five alternative methods for calculating mask size and demonstrate the advantages of our proposed method. Note that the CIoU reported in the table is CIoU@0.3.}
%   \vspace{-2mm}
  \label{sup-thres}
\end{table*}

\section{Details of Threshold for Binarized Mask}

Here, we discuss the details of threshold $\epsilon$ used to binarize the class-specific visual localization map $L^v$ and compare each alternatives both intuitively and quantitatively.

Given class-specific visual localization map $L^v\in\mathbb{R}^{K\times H\times W}$, it contains $K$ number of heatmaps $L^v_{1}, L^v_{2}, ..., L^v_{K}$, where $K$ is the number of total categories. To effectively eliminate the interference of off-screen noise, the Off-Screen Noise Filter should suppress the audio probability of those categories with low-response on class-specific visual localization map. Intuitively, we have the following five alternative options:

\begin{itemize}
    \item 1. Set the threshold as a constant value and calculate the audio-guided distribution $\text{p}^{av}$ by:
\begin{align}
    \text{p}^{av} &= \frac{\sum\limits_{h,w}(L^v \textgreater \epsilon_1) \circ \text{p}^{a}}{||\sum\limits_{h,w}(L^v \textgreater \epsilon_1) \circ \text{p}^{a}||_1}, \epsilon_1 = const.
    \label{sup-1}
\end{align}
    \item 2. Set a constant ratio $r$ and calculate the threshold by multiplying $r$ with the max response value of all class-specific visual localization maps $L^v$ in a batch. The audio-guided distribution $\text{p}^{av}$ is:
\begin{align}
    \text{p}^{av} &= \frac{\sum\limits_{h,w}(L^v \textgreater \epsilon_2) \circ \text{p}^{a}}{||\sum\limits_{h,w}(L^v \textgreater \epsilon_2) \circ \text{p}^{a}||_1}, \epsilon_2 = r \cdot \max\limits_{k,h,w}(L^v).
    \label{sup-2}
\end{align}
    \item 3. Set a constant ratio $r$ and calculate \textbf{the $i$-th} threshold by multiplying $r$ with the max response value of \textbf{the $i$-th} class-specific visual localization map $L^v$ ($i$ is the index in a batch during training stage). \textbf{The $i$-th} audio-guided distribution $\text{p}^{av}$ can be calculated by:
\begin{align}
    \text{p}^{av} &= \frac{\sum\limits_{h,w}(L^v \textgreater \epsilon_3) \circ \text{p}^{a}}{||\sum\limits_{h,w}(L^v \textgreater \epsilon_3) \circ \text{p}^{a}||_1}, \epsilon_3 = r \cdot \max\limits_{h,w}(L^v).
    \label{sup-3}
\end{align}
    \item 4. Instead of setting a certain threshold, we utilize the \textit{Global Average Pooling} (GAP) value of the $i$-th class-specific visual localization map $L^v$ to regularize the $i$-th audio-guided distribution $\text{p}^{av}$:
\begin{align}
    \text{p}^{av} &= \frac{\textit{GAP}(L^v) \circ \text{p}^{a}}{||\textit{GAP}(L^v) \circ \text{p}^{a}||_1}.
    \label{sup-4}
\end{align}
    \item 5. Set the threshold as the average response value of all class-specific visual localization maps $L^v$ in a batch. The audio-guided distribution $\text{p}^{av}$ is:
\begin{align}
    \text{p}^{av} &= \frac{\sum\limits_{h,w}(L^v \textgreater \epsilon_5) \circ \text{p}^{a}}{||\sum\limits_{h,w}(L^v \textgreater \epsilon_5) \circ \text{p}^{a}||_1},
    \label{sup-5_1}
\end{align}
where $\epsilon_5$ is calculated by:
\begin{align}
    \epsilon_5 = \frac{\sum\limits_{i=1}^B\sum\limits_{h=1}^H\sum\limits_{w=1}^W(L^v(h, w))}{B \cdot H \cdot W},
    \label{sup-5_2}
\end{align}
where $B$ denotes the batch size, $H$ denotes the height of feature map,  $W$ denotes the width of feature map and $L^v(h, w)$ is the response value of the $i$-th feature map with spatial row index $h$ and column index $w$. Note that we empirically set $B = 32$ in practice.
\end{itemize}

For the first choice of threshold, since the existence of objects in a feature map can mostly be captured by relative difference compared to other regions, it is not robust to set a constant value as a threshold. 

For the second choice of threshold, it is vulnerable to noise. An extremely high response value caused by noise will set an unreasonable threshold and it will mistakenly filters out those object-existed regions.

As for the third option, it can not effectively suppress off-screen noise. Since those feature maps without in-the-scene objects mostly remain low response on the whole scene, its max response value over all regions may be very small. Therefore, the threshold fails to filters out off-screen noise.

The fourth threshold is also vulnerable to noise. Since the \textit{GAP} value of the $i$-th object retrieval map $L^v$ is calculated as:
\begin{align}
    \textit{GAP}(L^v) = \frac{\sum\limits_{h=1}^H\sum\limits_{w=1}^W(L^v(h, w))}{H \cdot W}.
    \label{sup-6}
\end{align}
It can be severely affected by extremely high response value in the feature map.

The robustness of the final choice of threshold lies in two aspects: 1) The threshold is calculated from the whole batch samples, which can decrease the influence of extremely high response caused by noise; 2) The binarize operation can effectively eliminate the unbalanced value caused by noise in average calculation mentioned in \ref{sup-5_2}. For example, an overhigh response 0.8 caused by the outlier is equivalent to four responses of 0.2 by the fourth threshold, but the binarize operation can alleviate its weight.

To demonstrate the robustness and effectiveness of our method, we implement extensive experiments and quantitative results are recorded in Table~\ref{sup-thres}, which represents the superiority of the proposed threshold.

\begin{table*}[]
    \centering
    \begin{tabular}{c|c|c|c|c}
        \hline
        \multicolumn{5}{c}{Feature Extraction} \\
        \hline
         Stage & Audio Pathway & Audio Output & Visual Pathway & Visual Output \\
         \hline
         Backbone & ResNet+GMP & 512 & ResNet & 512$\times$14$\times$14 \\
         %\hline
         Transform & FC 512-128-128 & 128 & 1$\times$1 Conv 512-128-128 & 128$\times$14$\times$14 \\
         %\hline
         Normalize & L2-Norm (Replicate) & 128$\times$14$\times$14 & L2-Norm & 128$\times$14$\times$14 \\
         \hline
         \multicolumn{5}{c}{Audio-Visual Correspondence} \\
         \hline
         \multicolumn{2}{c|}{Operation} & \multicolumn{3}{c}{Output Size} \\
         \hline
         \multicolumn{2}{c|}{Cosine Similarity} & \multicolumn{3}{c}{1$\times$14$\times$14} \\
         %\hline
         \multicolumn{2}{c|}{1$\times$1 Conv} & \multicolumn{3}{c}{1$\times$14$\times$14} \\
         %\hline
         \multicolumn{2}{c|}{GMP} & \multicolumn{3}{c}{1} \\
         \hline
    \end{tabular}
    \caption{\textbf{Details of the model architecture.} The number ahead of Conv layer indecates the kernel size, and the number followed by FC/Conv layer indicates the change of channel size. Note that the replicate operation indicates that we repeat the audio feature vector 14$\times$14 times to match the image feature dimension.}
    \label{sup-architecture}
\end{table*}

\section{Details of the Model Architecture}

Overall, we use variants of ResNet-18~\cite{he2016deep} as audio and visual backbones to extract deep features. Note that the final \emph{Global Average Pooling} (GAP) and fully-connected (FC) layer are removed. The details of audio subnet, visual subnet and audio-visual correspondence block are listed below and shown in Table.~\ref{sup-architecture}.

\paragraph{Audio Subnet.} Considering that the audio input is the spectrogram with channel size 1, we set the input channel size of the first convolution layer 1. To focus on the foreground audio semantics, we employ \emph{Global Max Pooling} (GMP) to transform output audio feature map into 512-D feature vector, noted as $f^a \in \mathbb{R}^{C}$ in the paper.

\paragraph{Visual Subnet.} To maintain the fine-grained spatial information in visual scene, we set the stride in the last convolution block of ResNet-18 to 1 to keep high resolution. In this way, the $14\times14$ image feature map is used to calculate the audio-visual association map (localization map) $l(f^v, f^a) \in \mathbb{R}^{H\times W}$ as well as the class-aware audio-visual localization map $L^{av}$, which improves the visual scene analysis.

\paragraph{Audio-Visual Correspondence.} Following~\cite{arandjelovic2018objects}, we use two FC layers of 512-128-128 to reduce the dimension of audio and visual features. Then we calculate the cosine similarity between 128-D audio feature vector and 128-D visual feature on each spatial grid to generate audio-visual association map (localization map) representing the likelihood of sounding area. The global max pooled localization map is regarded as a predicted indicator to reveal whether the audio and visual message corresponds.

\begin{table*}[!h]
    \centering
        \begin{tabular}{cc|cc|cc|cc|cc}
            \hline
            \multicolumn{2}{c|}{Dataset} & \multicolumn{2}{c|}{MUSIC-Syn.} & \multicolumn{2}{c|}{MUSIC-Duet} & \multicolumn{2}{c|}{MUSIC-Un.} & \multicolumn{2}{c}{VGG-Un.} \\
            \hline
            Audio-Instance-Identifier & Cross-modal Referrer & CIoU & AUC & CIoU & AUC & CIoU & AUC & CIoU & AUC\\
            \hline
            \XSolidBrush & \XSolidBrush  & 0.2 & 6.8 & 6.7 & 12.1 & 0.9 & 1.2 & 3.1 & 7.8 \\
            \CheckmarkBold & \XSolidBrush  & 12.8 & 11.1 & 38.1 & 24.2 & 2.7 & 6.9 & 5.9 & 10.2\\
            \XSolidBrush & \CheckmarkBold  & 29.5 & 22.3 & 30.5 & 19.8 & 1.9 & 4.4 & 6.9 & 13.5\\
            \hline
            \CheckmarkBold & \CheckmarkBold  & \textbf{47.6} & \textbf{29.8} & \textbf{52.9} & \textbf{33.8} & \textbf{15.6} & \textbf{15.3} & \textbf{12.8} & \textbf{17.6}\\
            \hline
        \end{tabular}
        \vspace{-1mm}
        \caption{\textbf{Ablation study on Audio-Instance-Identifier and Cross-modal Referrer in terms of CIoU@0.3 and AUC for MUSIC-Syn., MUSIC-Duet, MUSIC-Un. and VGG-Un.} Note that we choose best prototype which achieves 0.809 NMI, 0.461 precision and 0.715 recall on MUSIC; 0.436 NMI, 0.346 precision and 0.232 recall on VGGSound for later training.}
        \vspace{0mm}
        \label{sup-tbl:whole}   
\end{table*}

\begin{table*}[!h]
    \centering
        \begin{tabular}{cc|cc|cc|cc|cc}
            \hline
            \multicolumn{2}{c|}{Dataset} & \multicolumn{2}{c|}{MUSIC-Syn.} & \multicolumn{2}{c|}{MUSIC-Duet} & \multicolumn{2}{c|}{MUSIC-Un.} & \multicolumn{2}{c}{VGG-Un.} \\
            \hline
            Silent Object Filter & Off-Screen Noise Filter & CIoU & AUC & CIoU & AUC & CIoU & AUC & CIoU & AUC\\
            \hline
            \XSolidBrush & \XSolidBrush  & 12.8 & 11.1 & 38.1 & 24.2 & 2.7 & 6.9 & 5.9 & 10.2 \\
            \CheckmarkBold & \XSolidBrush  & 37.1 & 25.2 & 44.7 & 26.7 & 4.6 & 9.5 & 7.2 & 14.1\\
            \XSolidBrush & \CheckmarkBold  & 11.4 & 13.2 & 43.8 & 25.4 & 5.1 & 10.8 & 7.8 & 13.9\\
            \hline
            \CheckmarkBold & \CheckmarkBold  & \textbf{47.6} & \textbf{29.8} & \textbf{52.9} & \textbf{33.8} & \textbf{15.6} & \textbf{15.3} & \textbf{12.8} & \textbf{17.6}\\
            \hline
        \end{tabular}
        \vspace{-1mm}
        \caption{\textbf{Ablation study on Silent Object Filter and Off-Screen Noise Filter in terms of CIoU@0.3 and AUC for MUSIC-Syn., MUSIC-Duet, MUSIC-Un. and VGG-Un.} Note that we choose best prototype which achieves 0.809 NMI, 0.461 precision and 0.715 recall on MUSIC; 0.436 NMI, 0.346 precision and 0.232 recall on VGGSound for later training.}
        \vspace{0mm}
        \label{sup-tbl:misalign}   
\end{table*}

\begin{table*}[!h]
    \centering
        \begin{tabular}{c|c|c|cc|cc|cc|cc}
            \hline
            \multicolumn{3}{c|}{Num. of$\backslash$Methods} & \multicolumn{2}{c|}{Object-that-Sound} & \multicolumn{2}{c|}{Sound-of-Pixel}  & \multicolumn{2}{c|}{DSOL}  & \multicolumn{2}{c}{\textbf{IEr (Ours)}} \\
            \hline
            sounding object & silent object& off-screen sound & CIoU & AUC & CIoU & AUC & CIoU & AUC & CIoU & AUC\\
            \hline
            2 & 2 & 1 & 0.1 & 6.8 & 7.5 & 11.6 & 3.2 & 7.3 & \textbf{15.6} & \textbf{15.3} \\
            3 & 3 & 2 & 0 & 3.6 & 3.1 & 6.7 & 1.3 & 4.8 & \textbf{14.7} & \textbf{14.2} \\
            4 & 4 & 3 & 0 & 1.9 & 1.4 & 4.2 & 0.5 & 2.7 & \textbf{12.8} & \textbf{12.6} \\
            
            \hline
        \end{tabular}
       \caption{\textbf{Ablation study on Robustness of Audio-Instance-Identifier (AII) in different SNRs.} Note that we combine different numbers of sounding objects, silent objects and off-screen  sounds to simulate different SNRs (interference levels).}
       \label{sup-tbl:dy}
\end{table*}

\section{Calculation of ${IoU}_t$ in CIoU Metric}

Each sample possesses a true label and a predicted pseudo label. Given a pseudo class, we correspond it to a true category according to its samples' most numbered true labels.

\section{Details of the Training Process}

We employ deep clustering on visual object features to generate $K$ cluster centroids as visual prototype $\mathcal{P}^v$ and $K$-dimensional one-hot cluster assignments $\mathcal{Y}^s$ indicating the pseudo class label of each single-sound sample $\mathcal{X}^s$ to guide classification training and generate audio prior $\mathcal{P}^a$. Motivated by~\cite{caron2018deep}, the number of cluster $K$ does not need to be the same as the number of category in training data, when $K$ is comparatively large, it significantly boosts the discrimination of extracted features. Therefore, to ensure that each category of sound source can be distinguished, we set $K$ larger than the real number of category in MUSIC/VGGSound. And in evaluation, we build to many-to-one mapping to allow multiple clusters to correspond to the same category for class-aware sound localization. In this way, the training process does not require the prior of specific number of categories in training data.

\paragraph{Curriculum Learning Strategy.}

To effectively enhance the model's ability to achieve mixed sound perception, we propose to train by an easy-to-hard manner. Specifically, we generate the mixed sound samples by averaging the waveforms of individual samples. Besides, we dynamically increase the ratio and difficulty of hard tasks by updating how many single-sound samples are mixed and how likely do we add mixed sound samples to training data respectively. In the training phase, we initialize $p = 0.5$ and gradually rises to $0.9$ according to its training progress in a batch. Similarly, the parameter of number of mixtures is initialized to be 2 and evenly increases to 4.

\section{Ablation Study}
In this section, we perform more ablation study experiments on each module proposed in the framework. Then we explore the robustness of Audio-Instance-Identifier (AII) in different SNRs (different interference levels).

\noindent\textbf{Audio-Instance-Identifier and Cross-modal Referrer.}
\\
To quantify the efficacy of Audio-Instance-Identifier and Cross-modal Referrer, we present the ablation study results in terms of CIoU and AUC for four mixed sound datasets in Table~\ref{sup-tbl:whole}. It indicates that Audio-Instance-Identifier and Cross-modal Referrer can both lead to improvement over discriminative sound source localization evaluation metrics. The significant improvement majorly comes from: 1) \textbf{Audio-Instance-Identifier} facilitates expanded class-specific audio representation that greatly improves mixed sound perception, while previous methods mostly ignore this step. 2) \textbf{Cross-modal Referrer} eliminates pervasive audio-visual interference of visible but silent objects and off-screen sounds in-the-wild and conducts category-level audio-visual matching, while previous methods only adopt coarse-grained correspondence or fail to filter out such interference.
\\
\noindent\textbf{Silent Object Filter and Off-Screen Noise Filter.}
\\
As shown in Table~\ref{sup-tbl:misalign}, both the Silent Object Filter and the Off-Screen Noise Filter can improve the model's localization ability on four datasets. Note that since the MUSIC-Duet dataset is almost free of off-screen noise and in-the-scene silent objects, the two filters have relatively low improvement on CIoU and AUC. Intuitively, the impact of the Silent Object Filter lies in its ability to suppress noise on the class-aware audio-visual localization maps $L^{av}$ and the impact of the Off-Screen Noise Filter lies in its latent benefit of size-irrelevant distribution alignment mentioned in the Section 3.3 in the paper. Specifically, the visual-guided distribution $p^{va}$ is highly unbalanced due to diverse object size, while the audio-guided distribution $p^{av}$ merely reflects sounding probability. For example, the duet of a big guitar and a slender flute will generate a balanced sound distribution with the help of our proposed Audio-Instance-Identifier. But the \textit{GAP} operation with softmax regression will give unreasonably high probability for guitar. And the aforementioned benefit of size-irrelevant alignment is also demonstrated in MUSIC-Synthetic, which do not contain off-screen noise. By observing the performance on MUSIC-Synthetic, MUSIC-Unconstrained and VGG-Unconstrained, we can conclude that Silent Object Filter and Off-Screen Noise Filter can effectively eliminate audio-visual interference in general scenes.
\\
\noindent\textbf{Robustness of Audio-Instance-Identifier (AII) in different SNRs.}
\\
We combine different numbers of sounding objects, silent objects and off-screen sounds to simulate different interference levels to compare the robustness of baseline methods and ours. The results are recorded in Table~\ref{sup-tbl:dy}. We can see that our IEr generalizes well to severe interference settings, performing consistently better than baselines. This can further verify that our proposed method is robust to in-the-wild scenarios with multiple interference levels.

\section{Additional Qualitative Results}

\begin{figure*}
    \centering
    \subfigure[Engine]{
    \includegraphics[width=0.117\linewidth]{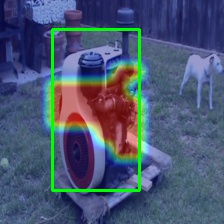}
    \includegraphics[width=0.117\linewidth]{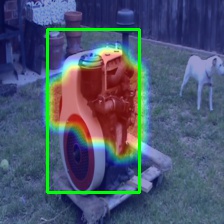}
    \includegraphics[width=0.117\linewidth]{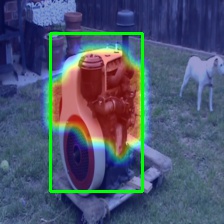}
    \includegraphics[width=0.117\linewidth]{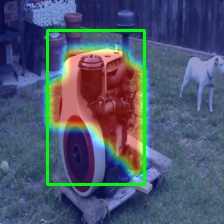}
    \includegraphics[width=0.117\linewidth]{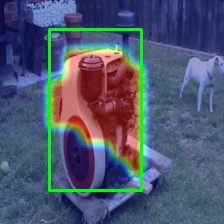}
    \includegraphics[width=0.117\linewidth]{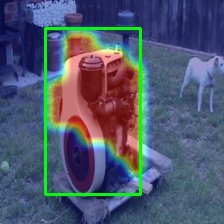}
    \includegraphics[width=0.117\linewidth]{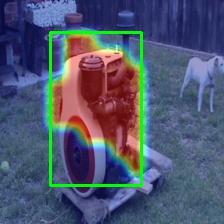}
    \includegraphics[width=0.117\linewidth]{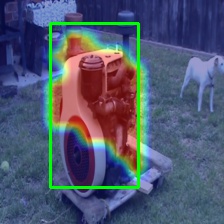}
    }
    \vspace{-2mm}\\
    \subfigure[Silent dog]{
    \includegraphics[width=0.117\linewidth]{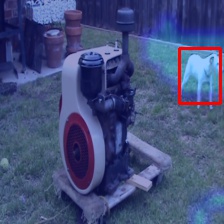}
    \includegraphics[width=0.117\linewidth]{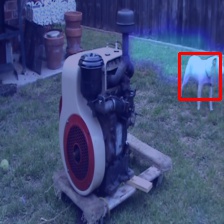}
    \includegraphics[width=0.117\linewidth]{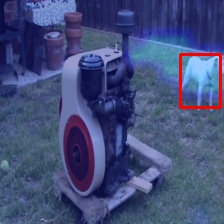}
    \includegraphics[width=0.117\linewidth]{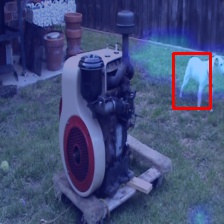}
    \includegraphics[width=0.117\linewidth]{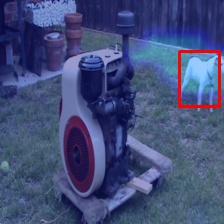}
    \includegraphics[width=0.117\linewidth]{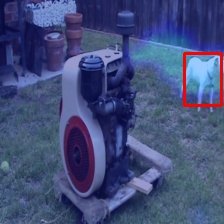}
    \includegraphics[width=0.117\linewidth]{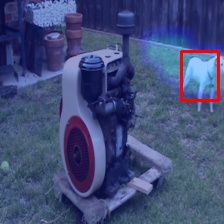}
    \includegraphics[width=0.117\linewidth]{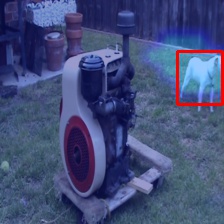}
    }
    \vspace{-2mm}
    \caption{\textbf{Engine with silent dog.} The first row is localization map of engine, the second row is localization map of silent dog, note that the activation value in silent area should be suppressed.}
    \label{sup-engine}
\end{figure*}

\begin{figure*}
    \centering
    \subfigure[Toy-car]{
    \includegraphics[width=0.117\linewidth]{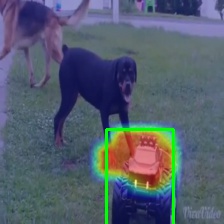}
    \includegraphics[width=0.117\linewidth]{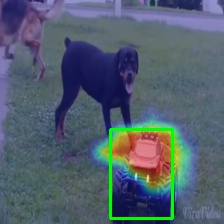}
    \includegraphics[width=0.117\linewidth]{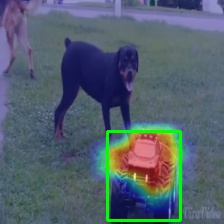}
    \includegraphics[width=0.117\linewidth]{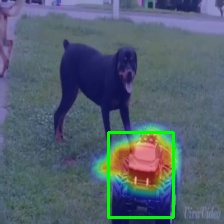}
    \includegraphics[width=0.117\linewidth]{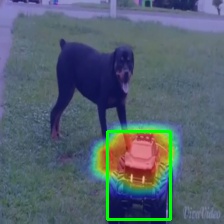}
    \includegraphics[width=0.117\linewidth]{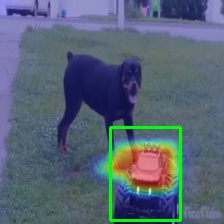}
    \includegraphics[width=0.117\linewidth]{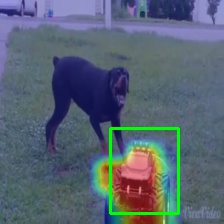}
    \includegraphics[width=0.117\linewidth]{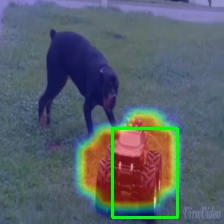}
    }
    \vspace{-2mm}\\
    \subfigure[Dog]{
    \includegraphics[width=0.117\linewidth]{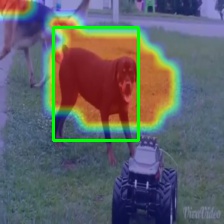}
    \includegraphics[width=0.117\linewidth]{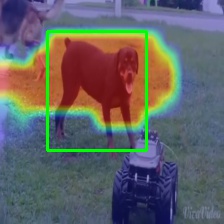}
    \includegraphics[width=0.117\linewidth]{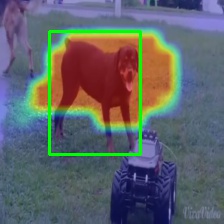}
    \includegraphics[width=0.117\linewidth]{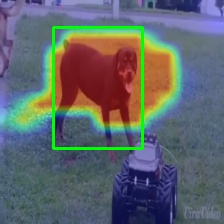}
    \includegraphics[width=0.117\linewidth]{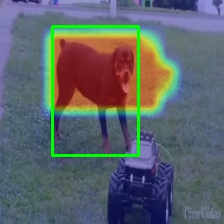}
    \includegraphics[width=0.117\linewidth]{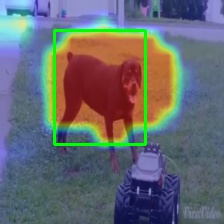}
    \includegraphics[width=0.117\linewidth]{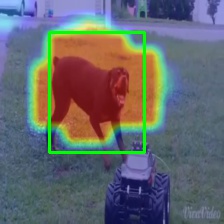}
    \includegraphics[width=0.117\linewidth]{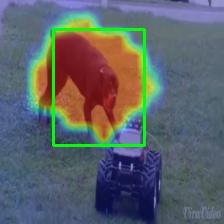}
    }
    \vspace{-2mm}
    \caption{\textbf{Toy-car and barking dog.} The first row is localization map of toy-car, the second row is localization map of dog.}
    \label{sup-toy}
\end{figure*}

\begin{figure*}
    \subfigure[Clarinet]{
    \includegraphics[width=0.117\linewidth]{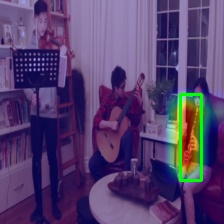}
    \includegraphics[width=0.117\linewidth]{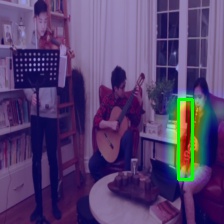}
    \includegraphics[width=0.117\linewidth]{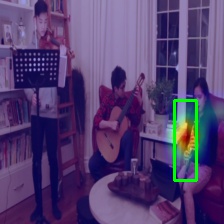}
    \includegraphics[width=0.117\linewidth]{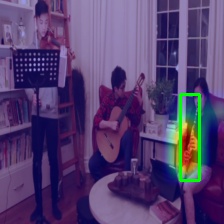}
    \includegraphics[width=0.117\linewidth]{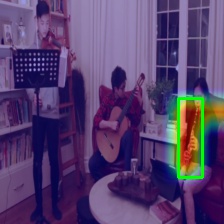}
    \includegraphics[width=0.117\linewidth]{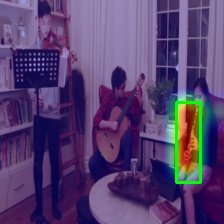}
    \includegraphics[width=0.117\linewidth]{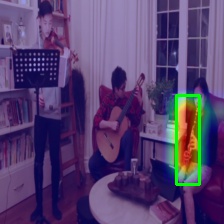}
    \includegraphics[width=0.117\linewidth]{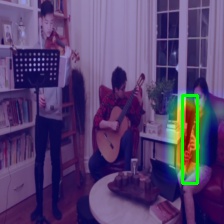}
    }
    \vspace{-2mm}\\
    \subfigure[Guitar]{
    \includegraphics[width=0.117\linewidth]{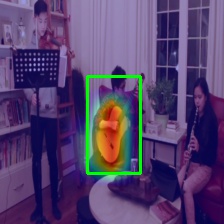}
    \includegraphics[width=0.117\linewidth]{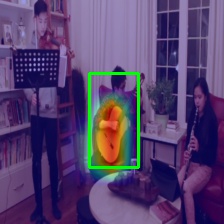}
    \includegraphics[width=0.117\linewidth]{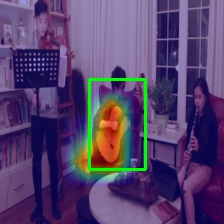}
    \includegraphics[width=0.117\linewidth]{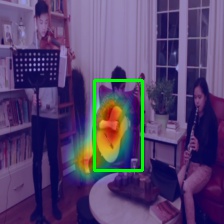}
    \includegraphics[width=0.117\linewidth]{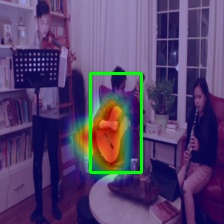}
    \includegraphics[width=0.117\linewidth]{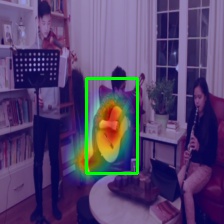}
    \includegraphics[width=0.117\linewidth]{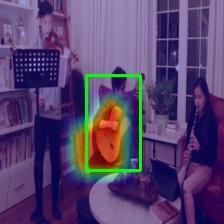}
    \includegraphics[width=0.117\linewidth]{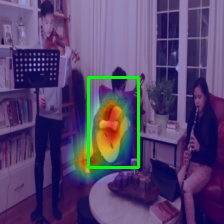}
    }
    \vspace{-2mm}\\
    \subfigure[Silent Violin]{
    \includegraphics[width=0.117\linewidth]{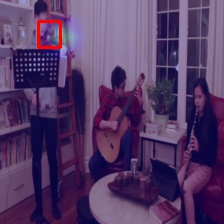}
    \includegraphics[width=0.117\linewidth]{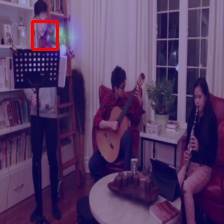}
    \includegraphics[width=0.117\linewidth]{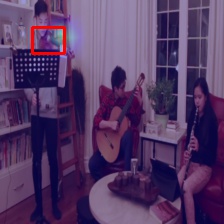}
    \includegraphics[width=0.117\linewidth]{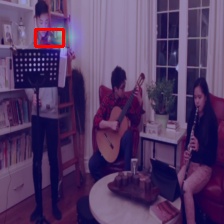}
    \includegraphics[width=0.117\linewidth]{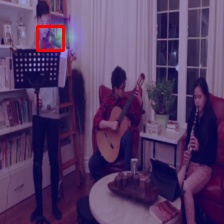}
    \includegraphics[width=0.117\linewidth]{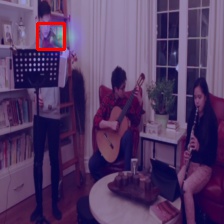}
    \includegraphics[width=0.117\linewidth]{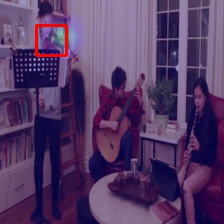}
    \includegraphics[width=0.117\linewidth]{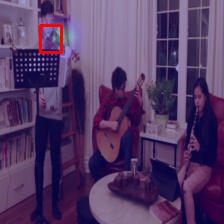}
    }
    \vspace{-2mm}
    \caption{\textbf{Clarinet and guitar with silent violin.} The first row is localization map of clarinet, the second row is localization map of guitar, the third row is localization map of silent violin, note that the activation value in silent area should be suppressed.}
    \label{sup-trio}
\end{figure*}

The additional qualitative results on discriminative sound source localization in-the-wild is shown below in Figure~\ref{sup-duet}, \ref{sup-machine}, \ref{sup-race}, \ref{sup-engine}, \ref{sup-toy}, \ref{sup-trio}. The green box in the localization map indicates the sounding object, the red box indicates the visible but silent one. In audio perception histogram, the green column indicates in-the-scene sound, the red column means in-the-scene silent category, and the blue column reveals the off-screen sound. The results show that our method can discriminatively localize objects and distinguish whether they are sound or silent. For audio perception, our method achieves category-balanced multi-sound audio perception, and can effectively suppress off-screen noise.

\section{Discussion and Furture Work}
To sum up, our \textbf{IEr} framework can \textit{semantically} associate sound-object pairs for in-the-wild audio-visual scenes. For example, in a concert scene, our method can attribute specific sounds to the corresponding category of instruments. However, when there are multiple visible objects of the same category, our model fails to determine the specific sounding instance. This is because we only input single static image and its corresponding audio clip without motion cues. It is quite difficult even for humans to determine the sounding individual among the multiple instances. In future work, we will try to integrate motion information to reach this goal. And for some typical applications in specific scenes like talking person identification, our model trained on general in-the-wild audio-visual scenes has difficulty achieving this goal, because the general categories are too coarse to discriminate specific individuals. But if we focus on the specific scenario, our framework could help to process mixed sound and erase the interferences for more robust audio-visual perception and sound localization.

\newpage

\begin{figure*}
    \subfigure[Guitar]{
    \includegraphics[width=0.117\linewidth]{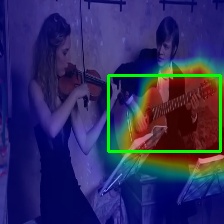}
    \includegraphics[width=0.117\linewidth]{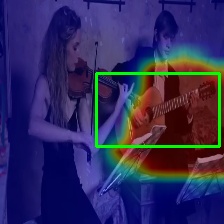}
    \includegraphics[width=0.117\linewidth]{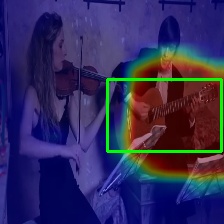}
    \includegraphics[width=0.117\linewidth]{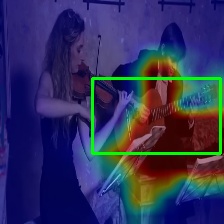}
    \includegraphics[width=0.117\linewidth]{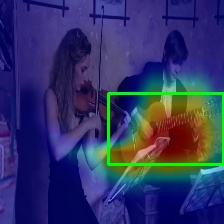}
    \includegraphics[width=0.117\linewidth]{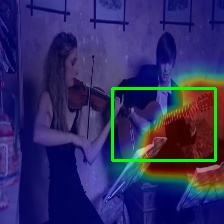}
    \includegraphics[width=0.117\linewidth]{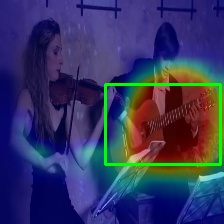}
    \includegraphics[width=0.117\linewidth]{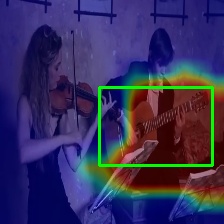}
    }
    \vspace{-2mm}\\
    \subfigure[Violin]{
    \includegraphics[width=0.117\linewidth]{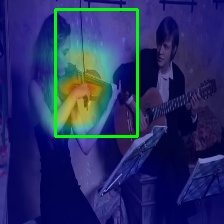}
    \includegraphics[width=0.117\linewidth]{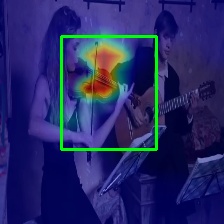}
    \includegraphics[width=0.117\linewidth]{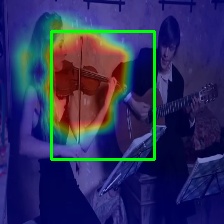}
    \includegraphics[width=0.117\linewidth]{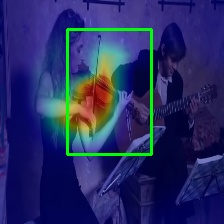}
    \includegraphics[width=0.117\linewidth]{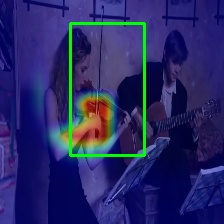}
    \includegraphics[width=0.117\linewidth]{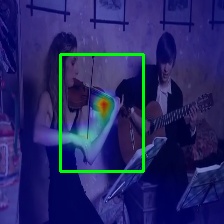}
    \includegraphics[width=0.117\linewidth]{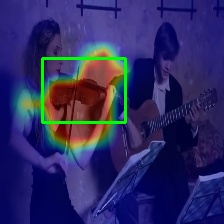}
    \includegraphics[width=0.117\linewidth]{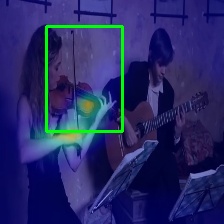}
    }
    \vspace{-2mm}
    \caption{\textbf{Duet of violin and guitar.} The first row is localization map of guitar, the second row is localization map of violin.}
    \label{sup-duet}
\end{figure*}

\begin{figure*}
    \centering
    \subfigure[Human Speech]{
    \includegraphics[width=0.117\linewidth]{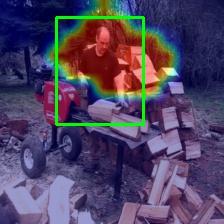}
    \includegraphics[width=0.117\linewidth]{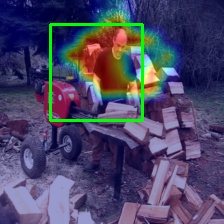}
    \includegraphics[width=0.117\linewidth]{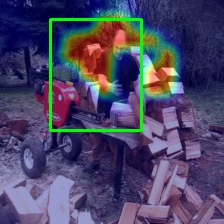}
    \includegraphics[width=0.117\linewidth]{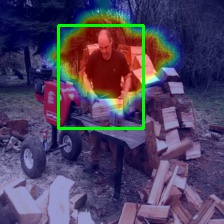}
    \includegraphics[width=0.117\linewidth]{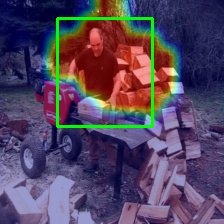}
    \includegraphics[width=0.117\linewidth]{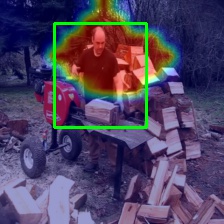}
    \includegraphics[width=0.117\linewidth]{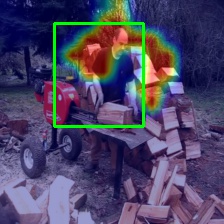}
    \includegraphics[width=0.117\linewidth]{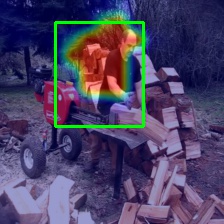}
    }
    \vspace{-2mm}\\
    \subfigure[Machine]{
    \includegraphics[width=0.117\linewidth]{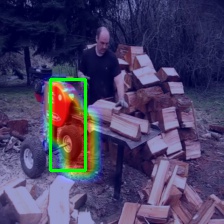}
    \includegraphics[width=0.117\linewidth]{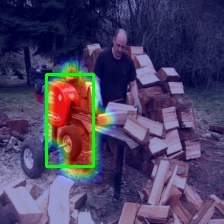}
    \includegraphics[width=0.117\linewidth]{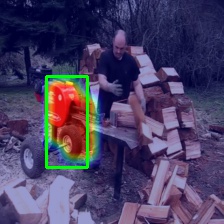}
    \includegraphics[width=0.117\linewidth]{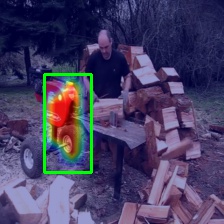}
    \includegraphics[width=0.117\linewidth]{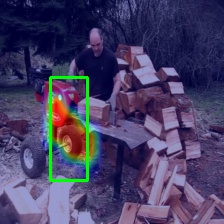}
    \includegraphics[width=0.117\linewidth]{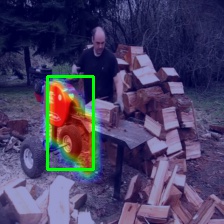}
    \includegraphics[width=0.117\linewidth]{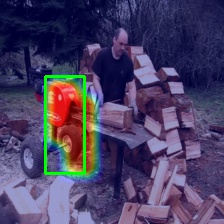}
    \includegraphics[width=0.117\linewidth]{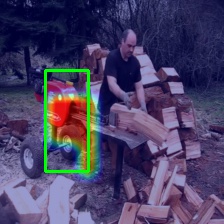}
    }
    \vspace{-2mm}
    \caption{\textbf{Machine and human speech.} The first row is localization map of the talking human, the second row is localization map of machine.}
    \label{sup-machine}
\end{figure*}

\begin{figure*}
    \subfigure[Race car]{
    \includegraphics[width=0.117\linewidth]{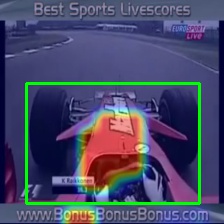}
    \includegraphics[width=0.117\linewidth]{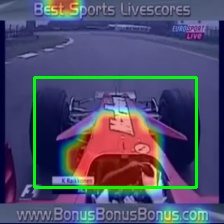}
    \includegraphics[width=0.117\linewidth]{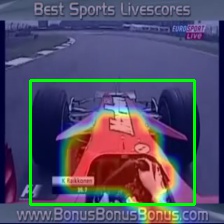}
    \includegraphics[width=0.117\linewidth]{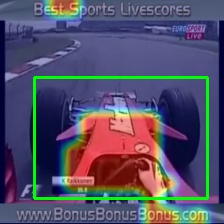}
    \includegraphics[width=0.117\linewidth]{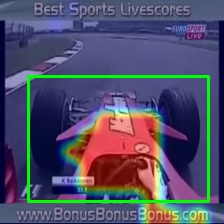}
    \includegraphics[width=0.117\linewidth]{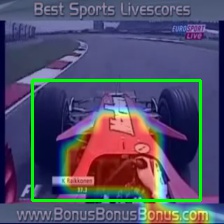}
    \includegraphics[width=0.117\linewidth]{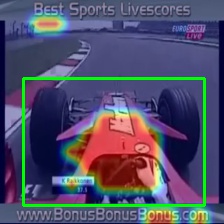}
    \includegraphics[width=0.117\linewidth]{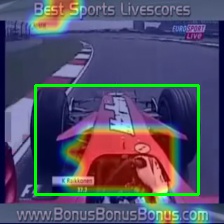}
    }
    \vspace{-2mm}\\
    \subfigure[All sound]{
    \includegraphics[width=0.117\linewidth]{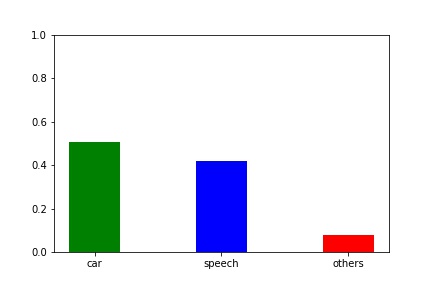}
    \includegraphics[width=0.117\linewidth]{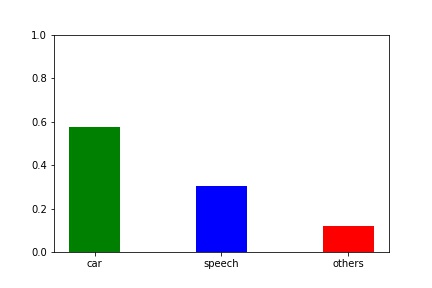}
    \includegraphics[width=0.117\linewidth]{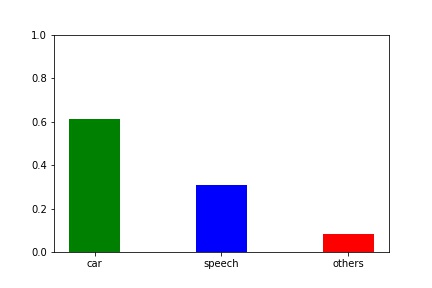}
    \includegraphics[width=0.117\linewidth]{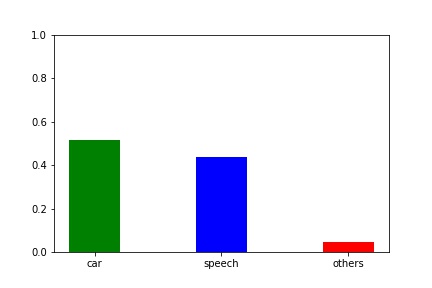}
    \includegraphics[width=0.117\linewidth]{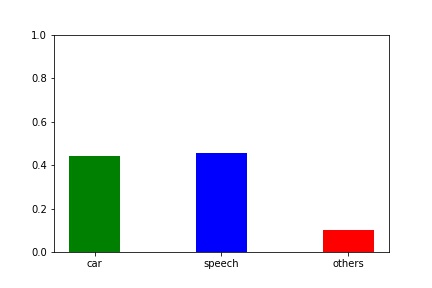}
    \includegraphics[width=0.117\linewidth]{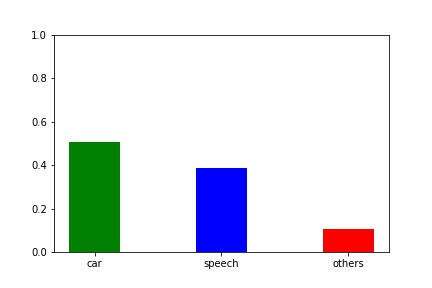}
    \includegraphics[width=0.117\linewidth]{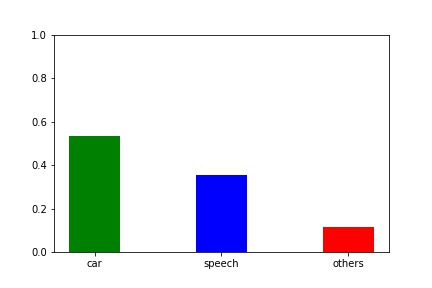}
    \includegraphics[width=0.117\linewidth]{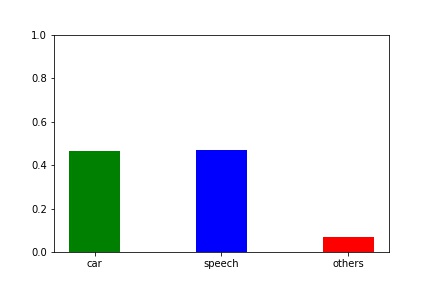}
    }
    \vspace{-2mm}\\
    \subfigure[In-the-scene sound]{
    \includegraphics[width=0.117\linewidth]{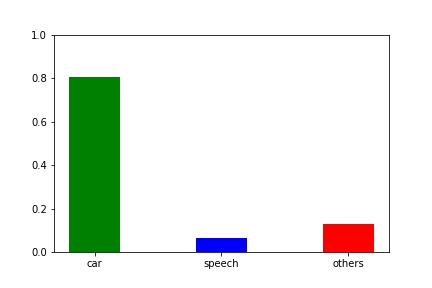}
    \includegraphics[width=0.117\linewidth]{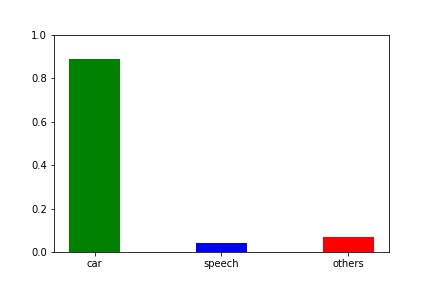}
    \includegraphics[width=0.117\linewidth]{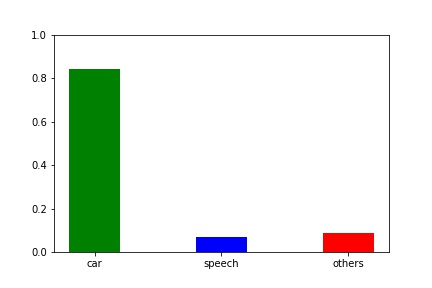}
    \includegraphics[width=0.117\linewidth]{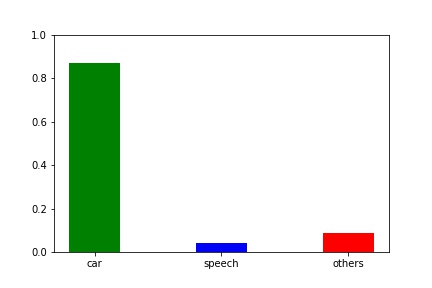}
    \includegraphics[width=0.117\linewidth]{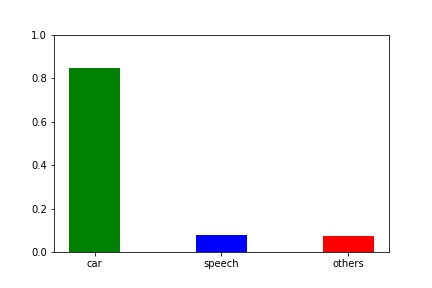}
    \includegraphics[width=0.117\linewidth]{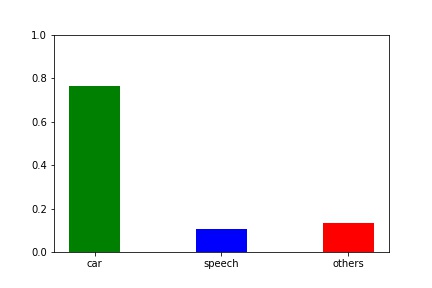}
    \includegraphics[width=0.117\linewidth]{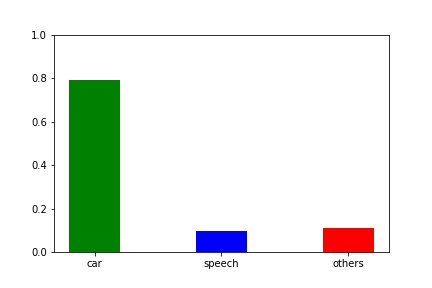}
    \includegraphics[width=0.117\linewidth]{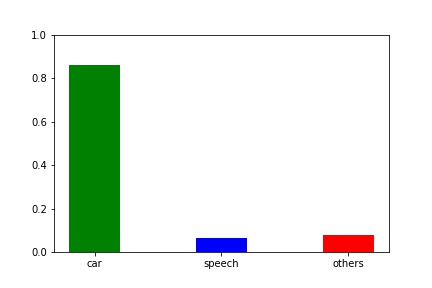}
    }
    \vspace{-2mm}
    \caption{\textbf{Race car with off-screen commentary.} The first row is localization map of race car, the second row is the histogram of overall sound perception, the third row is the histogram of in-the-scene sound perception.}
    \label{sup-race}
\end{figure*}

\end{document}